%% file: main.tex
\documentclass[11pt, a4paper]{article}
\usepackage{amsmath}
\usepackage{amsthm} 
\input{preamble}
\usepackage{xurl}  
\usepackage{array}
\newcommand{\file}{\nolinkurl}  

\newtheorem{theorem}{Theorem}

\newtheorem{lemma}{Lemma}
\newtheorem{proposition}{Proposition}
\newtheorem{corollary}{Corollary}
\theoremstyle{definition}
\newtheorem{remark}{Remark}
\crefname{theorem}{Theorem}{Theorems}
\crefname{lemma}{Lemma}{Lemmas}
\crefname{proposition}{Proposition}{Propositions}
\crefname{corollary}{Corollary}{Corollaries}
\crefname{remark}{Remark}{Remarks}

\newcommand{\R}{\mathbb{R}}
\newcommand{\C}{\mathbb{C}}
\newcommand{\F}{\mathcal{F}_{\mu,L}}
\newcommand{\Qc}{\mathcal{Q}_{\mu,L}}
\newcommand{\eps}{\varepsilon}
\newcommand{\re}{\operatorname{Re}}
\newcommand{\im}{\operatorname{Im}}
\newcommand{\acot}{\operatorname{arccot}}
\newcommand{\rest}{\mathrm{rest}}

\title{The Optimal Rate Constant of Heavy-Ball Momentum on Smooth Strongly Convex Functions}
\runningtitle{The Optimal Rate Constant of Heavy-Ball Momentum}
\author{Marcos Hernanz}
\contact{marcos@hernanz.io}
\headnote{\textbf{Preprint}}

\begin{document}

\maketitle

\begin{abstract}
  The heavy-ball method with a constant step size and a constant momentum parameter accelerates gradient descent on
  quadratics, but no choice of the two parameters does so on the class of smooth strongly convex functions. We determine how much faster
  than gradient descent it can be made on this class. Writing the best worst-case rate over all tunings as
  $1-c(\kappa)/\kappa$, where $\kappa$ is the condition number, we prove that $c(\kappa)$ tends to $8$ in every
  dimension of two or more. The lower bound on the rate holds for every tuning, because a tuning that is faster than
  $1-8/\kappa$ on quadratics follows a slowly decaying spiral on some function of two variables. The matching upper bound
  comes from an explicit tuning, with nearly the momentum of Polyak's tuning and a step size about $\sqrt\kappa/8$ times
  smaller, and from a Lyapunov function whose positivity is verified in exact arithmetic. Optimally tuned heavy-ball is
  therefore asymptotically four times faster than gradient descent on this class.
\end{abstract}

\tableofcontents

\section{Introduction}
\label{sec:introduction}

The heavy-ball method of \citet{polyak1964some} minimizes a differentiable function $f$ by the iteration
\begin{equation}
  x_{t+1}=x_t-\gamma\nabla f(x_t)+\beta(x_t-x_{t-1}),
  \label{eq:hb}
\end{equation}
with a step size $\gamma$ and a momentum parameter $\beta$. It is the prototype of the momentum methods used to train
neural networks. On quadratics whose Hessian has its eigenvalues in $[\mu,L]$, Polyak's choice of $\gamma$ and $\beta$
converges at the rate $1-2/\sqrt\kappa+O(1/\kappa)$, where $\kappa=L/\mu$ is the condition number. This is the
accelerated rate. When the dimension is large compared with the number of iterations, no first-order method has
a better one, already on quadratics \citep[Thm.~2.1.13]{nemirovski1983problem,nesterov2018lectures}. Gradient
descent, the case $\beta=0$, has the rate $1-2/\kappa+O(1/\kappa^2)$ with its best step size.

Beyond quadratics, heavy-ball loses this advantage. On the class $\F$ of $L$-smooth $\mu$-strongly convex functions,
Polyak's tuning fails to converge on some functions as soon as $\kappa\ge9+4\sqrt5$
\citep[Thm.~2.2]{pasechniuk2026two}, and \citet{goujaud2025provable} proved that no tuning accelerates in
dimension two or more. Every choice of $(\gamma,\beta)$ either fails to converge on some function of the class or
converges no faster than $1-O(1/\kappa)$, already on quadratics. Their counterexamples are cycles, found numerically by
\citet{goujaud2023counter} and then constructed in closed form. Non-acceleration holds in one dimension as well
\citep{fotopoulos2026cycling,liu2026sharp}.

This leaves a quantitative question. Let $\rho^\star(\kappa)$ be the best worst-case rate of~\eqref{eq:hb} on $\F$ over
all constant tunings $(\gamma,\beta)$, and write it as $\rho^\star(\kappa)=1-c(\kappa)/\kappa$. Gradient descent shows
that $c(\kappa)\ge2\kappa/(\kappa+1)$, and non-acceleration says that $c(\kappa)$ is bounded. How large is it?
\Citet[§7]{goujaud2025provable} note that the exact rate is open. The thesis of \citet[§10.3.3]{goujaud2024thesis}
records the proved bounds, which are asymptotically $2\le c\le100/3$ in our normalization, computes numerical optima at
$\kappa=20$, $100$ and $1000$, and asks what the limit is. It poses the related problem of finding analytical expressions
for the optimal tuning and rate. Concurrently with this work, \citet[App.~F]{liu2026sharp} proved a lower bound in one dimension that implies
$c\le24$ (\cref{sec:related}).

This paper determines the limit. In every dimension $d\ge2$ we prove that $c(\kappa)\to8$. The lower bound on the rate
(\cref{thm:lower}) holds for every tuning. If a tuning converges faster than $1-8/\kappa$ on quadratics, we construct a
function of two variables on which heavy-ball follows the spiral $x_t=(1-8/\kappa)^te^{2\pi it/K}$ around the
minimizer, for some integer $K$. The upper bound (\cref{thm:upper}) is an explicit tuning for $\kappa\ge400$, with
momentum $1-\beta\approx4/\sqrt\kappa$ and step size $\gamma\approx32/(L\sqrt\kappa)$, whose worst-case rate is
$1-(8-64/\sqrt\kappa)/\kappa$ in every dimension. Together they give $8-64/\sqrt\kappa\le c(\kappa)\le8$
(\cref{cor:optimal}). Optimally tuned heavy-ball is therefore asymptotically four times faster than gradient descent on
$\F$, and this factor cannot be improved. This answers the question of \citet{goujaud2024thesis} about the limit, and it
solves the problem posed there of the optimal tuning and rate to first order in $1/\kappa$.

The constant $8$ has a short explanation, which \cref{sec:intuition} gives in full. Along a spiral the gradient is the
iterate scaled and rotated by a fixed nonzero angle, which no quadratic can produce. Convexity allows such a rotation only in
exchange for curvature, and the exchange is most favorable at half the largest curvature $L$. Balancing the two gives a
condition $c\,(2w-1)\ge8w^2$ on a free parameter $w$, which some $w$ satisfies if and only if $c\ge8$. The tuning of
\cref{thm:upper} has the momentum of Polyak's tuning to first order and a step size about $\sqrt\kappa/8$ times smaller.
Its proof uses a Lyapunov function that perturbs the mechanical energy of the heavy ball. Three of its six coefficients
are forced by the quadratic on which the rate is attained. The other three were fitted numerically and then rounded, and
the positivity of the resulting quadratic form is proved for all $\kappa\ge400$ at once by exact root counting.

Several ingredients come from earlier work. Non-acceleration is due to \citet{goujaud2025provable} in dimension two and
higher, and to \citet{fotopoulos2026cycling} and \citet{liu2026sharp} in one dimension. Our spirals are the decaying
version of the root-of-unity cycles of \citet{goujaud2025provable} and the single-frequency case of the decaying waves of
\citet{pasechniuk2026two}, whose extra coordinate we use for the rest start. The tunings that beat
a given rate on quadratics are described in \citet[Lemma~2.4]{goujaud2025provable}. Our certificate belongs to the class
with one step of memory and the lag-one multiplier, which is the weighted off-by-one constraint of
\citet{lessard2016analysis} (\cref{sec:related}). At each fixed $\kappa>64.82$ the rate of our tuning can be derived from a
concurrent criterion of \citet[Thm.~7.2(b)]{pasechniuk2026two} (\cref{rem:pv}). What is new is the value $8$ of the
limit. The lower bound $1-8/\kappa$ for every tuning improves on $1-(100/3+o(1))/\kappa$ \citep{goujaud2025provable} and
on the concurrent $1-(24+o(1))/\kappa$ \citep{liu2026sharp}. The family that attains it to first order is explicit, and its certificate is
valid for all $\kappa\ge400$ at once.

Parts of both proofs are computer-assisted, and only exact arithmetic is used. Polynomial identities are verified
symbolically and the positivity of explicit univariate polynomials by exact Sturm sequences. \Cref{sec:computer} lists
every such step with its range, its method and its script. \Cref{sec:finite} compares the asymptotic result with what is
known at finite $\kappa$, including exact certificates at four values of $\kappa$. \Cref{app:onedim} collects results in
one dimension. There we prove that the constant $c_1(\kappa)$ of $\F(\R)$ is at most $50/3+o(1)$, show that the cycle
method used in earlier lower bounds cannot go below $16$, and give an exact non-convergent orbit of Polyak's tuning
started at rest, for every $\kappa\ge100$.

\section{Related Work}
\label{sec:related}

Global analyses of heavy-ball on $\F$ give no explicit rate better than that of gradient descent.
\Citet{ghadimi2015global} prove global linear convergence on an explicit region of tunings. Their rate is not explicit
in general, and on a sub-family it is the gradient-descent factor $1-\gamma\mu$ for function values. \Citet[Lemma~2.5]{goujaud2025provable}
show that the best rate over that region, even on quadratics, is $1-8/\kappa+o(1/\kappa)$. This constant $8$ is related
to ours but describes different tunings (\cref{sec:ghadimi}). Other global analyses of heavy-ball on $\F$, or of
momentum families that contain it, certify explicit rates for small steps $\gamma=O(\mu/L^2)$
\citep{shi2022understanding,can2022entropic}. \Citet{danilova2018non} study its non-monotone transient behavior on
quadratics and prove global convergence under the Polyak--{\L}ojasiewicz condition.
Polyak's own analysis is local \citep{polyak1964some}. \Citet{ugrinovskii2022global} prove global asymptotic
convergence, without a global rate, on a class of functions with sector-bounded gradients and optimize the local rate.
\Citet[§6]{sanzserna2021connections} show that heavy-ball has no Lyapunov function of the kind they construct for
Nesterov's method, which would certify an accelerated rate.

Polyak's tuning can fail on $\F$. \Citet{lessard2016analysis} found a one-dimensional function at $\kappa=25$ on
which its iterates approach a cycle. \Citet{badithela2019analysis} solved the linear matrix inequality of the weighted
off-by-one constraint analytically for Polyak's tuning and constructed a function on which the method enters a limit
cycle. As \citet[§1.1]{pasechniuk2026two} describe it, the bound reaches $1$ at $\kappa=9+4\sqrt5$, where a function of
one variable has a three-cycle. \Citet[Thm.~2.2]{pasechniuk2026two}
prove that $9+4\sqrt5$ is the exact threshold, with convergence on all of $\F$ below it and a non-convergent trajectory
at and above it. \Citet{wang2022provable} claim acceleration for a class of
Polyak--{\L}ojasiewicz functions. Their erratum (arXiv v3, App.~J) adds a co-diagonalization condition, and
\citet{goujaud2025riddles} point to an incorrect inequality in the remaining one-dimensional argument. With time-varying
parameters on smooth convex functions, \citet{ma2026lower} prove a lower bound for heavy-ball that holds for every
schedule fixed in advance, and \citet{he2026heavy} use randomized schedules to obtain a faster rate than the best known
one for gradient descent. We only consider constant
parameters.

Other lower bounds for every tuning come from cycles. \Citet{goujaud2025provable} constructed heavy-ball cycles on the
$K$-th roots of unity in the plane (their Thm.~3.5) and proved that every tuning outside the cycling regions satisfies
$\rho\ge(1-C/\kappa)/(1+C/\kappa)$ for every $C>50/3$ (Thm.~3.6), which is $c\le100/3$ in our normalization.
\Citet{goujaud2025riddles} asked whether heavy-ball accelerates on functions of one variable.
\Citet{fotopoulos2026cycling} answered this negatively by rotating the planar cycles and projecting them onto a line.
Building on their construction, \citet[App.~F]{liu2026sharp} prove that in one dimension every tuning with $\gamma>0$
and $0\le\beta<1$ has rate at least $(\kappa-12)/(\kappa+12)$ when $\kappa\ge24$. Tunings outside this range have
$\rho_Q\ge1-8/\kappa$ by \cref{lem:jury}, which is larger. Since $\rho^\star$ does not decrease with the dimension, this gives $c\le24$ in every dimension.

The concurrent work of \citet{pasechniuk2026two} is the closest to ours. It answers the second riddle of
\citet{goujaud2025riddles} and develops a rate theory for general tunings, for the rest-start rate taken uniformly over the
dimension. Interpolable mixtures of rotating geometric sequences, decaying ones included, give lower bounds (their
Lemma~3.4 and Thm.~3.5(c)), and a lag-one Zames--Falb multiplier gives upper bounds (Thm.~7.2(a)). On an explicit region
$Q$ of tunings the rate equals the rate on quadratics (Thm.~7.2(b)). They do not optimize over tunings and do not state a limiting constant for the optimal rate. Among the
questions they leave open are the rates on the part of their region $\Omega_1$ outside $Q$ and the asymptotics of their
gaps as $\kappa\to\infty$ (their §10). The spiral of \cref{thm:lower} is the single-frequency case of their waves, and
the tuning of \cref{thm:upper} lies in their region $Q$ (\cref{rem:pv}). Our contribution relative to theirs is the
quantitative bound for every tuning and a family, uniform in $\kappa$, that matches it.

Worst-case rates for fixed parameters can be computed or certified by performance estimation
\citep{drori2014performance,taylor2017smooth}, by Lyapunov searches \citep{taylor2018lyapunov,upadhyaya2023automated}
and by integral quadratic constraints \citep{megretski1997system,lessard2016analysis}. \Citet{lessard2020direct}
synthesize algorithms with bounds on the achievable rate. For heavy-ball with step $1/L$ and numerically optimized
momentum, \citet[Fig.~8]{lessard2016analysis} report rates between about $1-6.4/\kappa$ and $1-7.1/\kappa$ for
$\kappa$ between about $850$ and $4200$ (our reading of their figure, whose resolution we discuss in
\cref{sec:earlier}). They describe this as comparable to gradient
descent, which is true of the order in $\kappa$. \Citet[Fig.~4]{vanscoy2021speed} find $c\approx5$ at $\kappa=100$, and
\citet[Fig.~2c]{upadhyaya2023automated} find $c\approx2.2$ at $\kappa=10$ with step $1/L$, both read off their figures and
converted from rates of the squared distance. Goujaud's thesis \citep[Table~10.1]{goujaud2024thesis} reports optima at $\kappa=20$, $100$ and
$1000$ (\cref{sec:earlier}). The lag-one multiplier behind our certificate is a discrete-time analogue of the Zames--Falb multipliers
\citep{zames1968stability}, and
\citet[§1.5]{pasechniuk2026two} note that it coincides with the weighted off-by-one constraint of
\citet{lessard2016analysis}. They relate single-harmonic periodic orbits to the periodic counterexamples in absolute
stability of \citet{seiler2021construction}, and multi-harmonic obstructions to the duality bounds for Zames--Falb
multipliers of \citet{zhang2022duality}.

Spirals have been used for Nesterov's method as well. \Citet{pokutta2026exact} determine the exact worst-case asymptotic rate
of Nesterov's method with its standard parameters, using planar spirals with a common contraction factor and a linear
program over measures. That work is concurrent and concerns a different method, but its tools are close to ours.

Methods that evaluate the gradient elsewhere do accelerate on $\F$. Measured, as here, by the distance to the
minimizer, Nesterov's method \citep{nesterov1983method} with constant momentum converges at rate at most
$(1-1/\sqrt\kappa)^{1/2}\approx1-1/(2\sqrt\kappa)$ \citep[§2.2.1]{nesterov2018lectures}, and the triple momentum method at
rate $1-1/\sqrt\kappa$ \citep{vanscoy2018fastest}. No first-order method has a better rate for the distance to the
minimizer when the dimension is large compared with the number of iterations. On $\F(\R^{2N+1})$ the lower bound of \citet{drori2022oracle} after $N$ steps is of order
$(1-1/\sqrt\kappa)^N$, and the method of \citet{taylor2023optimal} attains it. Nesterov's method and the triple momentum
method evaluate the gradient at a point other than the iterate, which heavy-ball does not. If $f$ is in addition twice
continuously differentiable near the minimizer, the asymptotic rate of a globally convergent method is its local rate,
which for heavy-ball can be the accelerated rate of quadratics \citep{polyak1964some}. \Citet{ugrinovskii2022global}
exploit this, and so do \citet{vanscoy2025fastest} with a new method. The functions in our lower bound are not twice
differentiable at the minimizer, and cannot be. If $f$ were twice differentiable at $x_\star$, the iterates of
\cref{thm:lower} would eventually converge at a rate at most $\rho_Q<1-8/\kappa$.

\section{Setting and Main Results}
\label{sec:setting}

\subsection{Worst-case rates}

For $0<\mu<L$, let $\F(\R^d)$ be the set of differentiable functions $f:\R^d\to\R$ such that
$f-(\mu/2)\|\cdot\|^2$ is convex and $\nabla f$ is $L$-Lipschitz, and let $\Qc$ be its subset of quadratics. Each
$f\in\F$ has a unique minimizer $x_\star$. Heavy-ball~\eqref{eq:hb} starts from an arbitrary pair $(x_{-1},x_0)$, as in
the cycle constructions of \citet{goujaud2025provable}. For a tuning $(\gamma,\beta)\in\R^2$ and a set of functions
$\mathcal F$, the rate $\rho_{\gamma,\beta}(\mathcal F)$ is the smallest $\rho\ge0$ with the following property. For
every $\eps>0$ there is a $T_0$ such that
\begin{equation}
  \|x_T-x_\star\|\le(\rho+\eps)^T\max\bigl\{\|x_0-x_\star\|,\|x_{-1}-x_\star\|\bigr\}
  \label{eq:rate}
\end{equation}
for all $f\in\mathcal F$, all initial pairs and all $T\ge T_0$. The rest-start rate $\rho^\rest_{\gamma,\beta}$ is
defined in the same way with $x_{-1}=x_0$. The optimal rate in dimension $d$ and its constant are
\[
  \rho^\star_d(\kappa)=\inf_{(\gamma,\beta)\in\R^2}\rho_{\gamma,\beta}\bigl(\F(\R^d)\bigr),\qquad
  c_d(\kappa)=\kappa\bigl(1-\rho^\star_d(\kappa)\bigr).
\]
By scaling, $\rho^\star_d$ depends on $\mu$ and $L$ only through $\kappa=L/\mu$. It does not decrease with $d$, because
adding coordinates on which $f$ is $(\mu/2)\|\cdot\|^2$ embeds $\F(\R^d)$ into $\F(\R^{d+1})$. Gradient descent with
$\gamma=2/(L+\mu)$ has rate $(\kappa-1)/(\kappa+1)$ on $\F$ \citep[Thm.~2.1.15]{nesterov2018lectures}, so
$c_d(\kappa)\ge2\kappa/(\kappa+1)$.

\begin{remark}[conventions]
  \label{rem:conventions}
  \citet{goujaud2025provable} and \citet{goujaud2024thesis} write $\kappa$ for $\mu/L$ and state rates as
  $(1-C\mu/L)/(1+C\mu/L)$. In our notation this is $c=2C\kappa/(\kappa+C)\to2C$. Asymptotically, our constant
  $8$ is $C=4$ there, and their bound for every $C>50/3$ is $c\le100/3$. They normalize by $\|x_0-x_\star\|$ and do not
  specify the second initial point \citep[Defs.~1.3--1.5]{goujaud2025provable}. \Citet{pasechniuk2026two} use the rest
  start (their eq.~(1.1)) and take the supremum over all dimensions inside the rate (their eq.~(1.2)). Our results hold under each of
  these conventions. The lower bound holds for arbitrary initial pairs in $d\ge2$ and from rest in $d\ge3$, normalized
  by either initial distance. The upper bound holds for every initial pair in every dimension, with a constant
  independent of $d$. Normalized by $\|x_0-x_\star\|$ alone, it needs $x_{-1}$ to be determined by $x_0$, as in the rest
  start.
\end{remark}

\subsection{Notation and interpolation conditions}

From now on we take $L=1$ and write
\[
  q=\mu=1/\kappa,\qquad s=\sqrt q=\kappa^{-1/2},\qquad h=\gamma L,
\]
with $x_\star=0$ and $f(x_\star)=0$. On the quadratic $(\lambda/2)\|x\|^2$, heavy-ball is a linear recursion with
characteristic polynomial $z^2-(1+\beta-h\lambda)z+\beta$. Let $r(\lambda)$ be the largest modulus of its roots and
$\rho_Q=\max_{\lambda\in[q,1]}r(\lambda)=\max\{r(q),r(1)\}$. Running heavy-ball from an eigenvector, or from the real part
of a complex one, shows that $\rho_{\gamma,\beta}(\F)\ge\rho_Q$, and $\rho_Q$ is the rate on $\Qc$
\citep[Prop.~2.1]{goujaud2025provable}.

Both proofs rest on the interpolation conditions for $\F$. A finite set of triples $(x_i,g_i,f_i)$ is
\emph{$\F$-interpolable} if some $f\in\F(\R^d)$ has $f(x_i)=f_i$ and $\nabla f(x_i)=g_i$ for all $i$. By
\citet[Thm.~4]{taylor2017smooth}, this holds if and only if, for all $i$ and $j$,
\begin{equation}
  I(i,j):=f_i-f_j-\langle g_j,x_i-x_j\rangle
  -\frac{\|g_i-g_j\|^2+q\|x_i-x_j\|^2-2q\langle g_i-g_j,x_i-x_j\rangle}{2(1-q)}\ \ge\ 0.
  \label{eq:thg}
\end{equation}
The lower bound constructs data that satisfy these inequalities, and the upper bound adds two of them with non-negative
multipliers.

\subsection{Main results}

\begin{theorem}[lower bound]
  \label{thm:lower}
  Let $0<\mu<L$ and $d\ge2$. For every $(\gamma,\beta)\in\R^2$,
  \[
    \rho_{\gamma,\beta}\bigl(\F(\R^d)\bigr)\ \ge\ 1-\frac{8}{\kappa}.
  \]
  More precisely, let $\kappa>8$ and $r=1-8/\kappa$. Either $\rho_Q\ge r$, or there are an integer $K\ge4$, a function
  $f\in\F(\R^2)$ and an initial pair on which heavy-ball satisfies $x_t-x_\star=r^t\,R_{2\pi t/K}(x_0-x_\star)$ for all
  $t\ge0$, where $R_\theta$ is the rotation of the plane by the angle $\theta$.
\end{theorem}

\begin{theorem}[achievability]
  \label{thm:upper}
  Let $\kappa\ge400$, $s=\kappa^{-1/2}$, $C=8-64s$, and
  \begin{equation}
    r=1-Cs^2,\qquad \beta=r(r-4s),\qquad \gamma=\frac{Cs(4+Cs)}{L}.
    \label{eq:tuning}
  \end{equation}
  There is a constant $M$, depending only on $\kappa$, such that for every $d$, every $f\in\F(\R^d)$ and every initial
  pair $(x_{-1},x_0)$,
  \[
    \|x_t-x_\star\|\le M\,r^t\,\bigl(\|x_0-x_\star\|+\|x_{-1}-x_\star\|\bigr)\qquad\text{for all }t\ge0.
  \]
  The rate is attained on $f=(\mu/2)\|x\|^2$, already from rest. Hence
  \[
    \rho_{\gamma,\beta}\bigl(\F(\R^d)\bigr)=\rho^\rest_{\gamma,\beta}\bigl(\F(\R^d)\bigr)=1-\frac{8-64/\sqrt\kappa}{\kappa}.
  \]
\end{theorem}

The constant $M$ is given in the proof. For example, $M<78$ at $\kappa=400$ and $M<559$ at $\kappa=10^4$
(\cref{tab:Mconst}). The same certificate works for every
$\kappa\ge357.2901\ldots$ and for no smaller $\kappa$ (\cref{sec:range}).

\begin{corollary}[the optimal constant]
  \label{cor:optimal}
  For $d\ge2$ and $\kappa\ge400$,
  \[
    1-\frac8\kappa\ \le\ \rho^\star_d(\kappa)\ \le\ 1-\frac8\kappa+\frac{64}{\kappa^{3/2}},
    \qquad\text{that is,}\qquad
    8-\frac{64}{\sqrt\kappa}\ \le\ c_d(\kappa)\ \le\ 8.
  \]
  The upper bound $c_d(\kappa)\le8$ holds for all $\kappa>1$. In particular $c_d(\kappa)\to8$, while the constant
  $2\kappa/(\kappa+1)$ of gradient descent tends to $2$. In one dimension, $c_1(\kappa)\ge8-64/\sqrt\kappa$ for
  $\kappa\ge400$.
\end{corollary}

\begin{corollary}[rest start]
  \label{cor:rest}
  For $d\ge3$ and every $(\gamma,\beta)\in\R^2$, $\rho^\rest_{\gamma,\beta}(\F(\R^d))\ge1-8/\kappa$. Hence
  \cref{cor:optimal} holds for the rest-start rate as well, for $d\ge3$.
\end{corollary}

In terms of iterations, the spiral of \cref{thm:lower} has
$\|x_N-x_\star\|=r^{N+1}\max\{\|x_0-x_\star\|,\|x_{-1}-x_\star\|\}$. If $\rho_Q\ge r$, the trajectory $x_t=z^{t+1}$ on
$(\lambda/2)\|x\|^2$, with $\R^2$ identified with $\mathbb C$ and $z$ a root of modulus $\rho_Q$ for $\lambda=q$ or
$\lambda=1$, decays no faster. To reduce the distance to the minimizer by a factor $\eps<1$ on every function of
$\F(\R^2)$, every tuning therefore needs at least $\log(1/\eps)/\log(1/r)-1\ge(\kappa/8-1)\log(1/\eps)-1$ iterations.
The tuning of \cref{thm:upper} needs at most $\bigl\lceil\kappa\bigl(\log(1/\eps)+\log 2M\bigr)/(8-64/\sqrt\kappa)\bigr\rceil$.
Gradient descent needs about $(\kappa/2)\log(1/\eps)$ iterations, and the triple momentum method about
$\sqrt\kappa\log(1/\eps)$.

\begin{remark}[relation to Pasechniuk-Vilensky]
  \label{rem:pv}
  The tuning~\eqref{eq:tuning} lies in the region $Q$ of \citet[Thm.~7.2(b)]{pasechniuk2026two} for every
  $\kappa>64.82$. For this family, their membership conditions (Cor.~7.3(a),(c)) reduce to the positivity of univariate
  polynomials in $s$, which we checked by exact root counting (\cref{sec:computer}). On $Q$ they prove that the rate
  equals the rate on quadratics, which here is $r$. At each fixed $\kappa$, their theorem therefore gives the rate equality of
  \cref{thm:upper} independently. Their statement is for the rest start and loses an arbitrary $\eps$ in the
  rate, although their proof controls arbitrary initial states. \Cref{thm:upper} adds a short, self-contained
  certificate in the time domain with an explicit constant $M$ for arbitrary initial pairs, and it identifies a family
  that is asymptotically optimal by \cref{thm:lower}.
\end{remark}

\section{Where the Constant 8 Comes From}
\label{sec:intuition}

This section explains the two theorems without full proofs. We keep $L=1$.

\subsection{What quadratics allow}
\label{sec:quadratics}

On a quadratic of curvature $\lambda$, heavy-ball is a linear recursion. In the flattest direction, $\lambda=q$, it
converges at a real rate $r$ if $r$ is a root of $z^2-(1+\beta-hq)z+\beta$. Evaluating this polynomial at $z=1$ gives
$hq=(1-r)(1-\beta/r)$. A real rate $r=1-c/\kappa$ therefore requires
\[
  \frac{h}{1-\beta/r}=c.
\]
We call the left side the \emph{effective step}. For gradient descent it equals $h$, and stability in the stiffest
direction, $\lambda=1$, requires $h<2$, hence $c<2$. Momentum separates the two requirements. The stiff direction only
needs $h\lesssim2(1+\beta)$, while the effective step grows without bound as $\beta/r\to1$. The momentum cannot exceed
$r^2$, because the roots for $\lambda=q$ have product $\beta$, so one of them has modulus at least $\sqrt\beta$. In the
coordinates
\[
  X=1-\beta/r\qquad\text{and}\qquad h=\gamma L,
\]
the tunings that beat $r=1-8/\kappa$ on quadratics form a triangle (\cref{lem:jury};
\citealp[Lemma~2.4]{goujaud2025provable}). Its three sides are $X>\eta:=1-r$, which bounds the momentum, $h>8X$, which
says that the effective step exceeds $8$, and $h<(2-\eta)(2-X)$, which keeps the stiff direction stable. On quadratics
alone, $c$ can be as large as about $2\sqrt\kappa$, which Polyak's tuning attains. Everything that limits $c$ to $8$
comes from functions that are not quadratic.

\subsection{Spirals}
\label{sec:why8}

\begin{figure}[t]
  \centering
  \includegraphics[width=\linewidth]{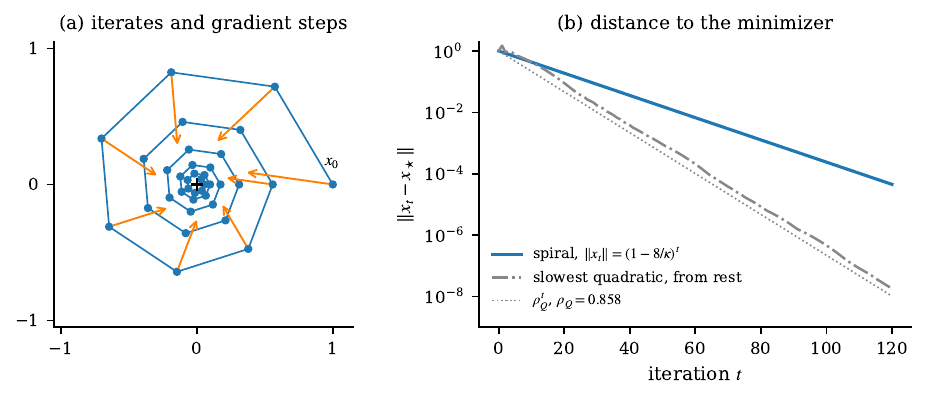}
  \caption{A spiral of \cref{lem:spiral} at $\kappa=100$, for the tuning $\gamma L=2.5$, $\beta=0.736$. (a)~The iterates
    turn by $2\pi/7$ per step and contract by $r=1-8/\kappa=0.92$. The gradients are the iterates scaled and rotated by a
    fixed nonzero angle, which no quadratic produces. The orange arrows are the gradient steps $-\gamma\nabla f(x_t)$. (b)~On this function heavy-ball converges at
    the rate $0.92$, although on every quadratic of the class it converges at the rate $\rho_Q=0.858$ or faster.}
  \label{fig:spiral}
\end{figure}

Identify $\R^2$ with $\C$ and look for a heavy-ball trajectory that spirals into the minimizer, $x_t=r^te^{i\theta t}$.
The iteration~\eqref{eq:hb} determines the gradients along it, $g_t=(a+ib)\,x_t$ with real $a$ and $b$ that depend on the
tuning, by~\eqref{eq:A}. The part $a\,x_t$ is what the quadratic $(a/2)\|x\|^2$ would produce. The part $ib\,x_t$ is
perpendicular to $x_t$ (\cref{fig:spiral}). No quadratic produces it along the whole spiral, because a Hessian $H$ with
$Hx_t=(a+ib)x_t$ for all $t$ is the multiplication by $a+ib$, which is symmetric only if $b=0$. A convex function that is
not quadratic can produce some of this rotation, but only a limited amount. The interpolation conditions allow it as long as
\[
  |b|\ \le\ 2m\tan\frac\theta2,\qquad\text{where}\qquad m=\frac{(a-q)(1-a)-b^2}{2(1-q)}\approx\frac{a(1-a)}2
\]
(\cref{lem:spiral}). The rotation $b$ has to be paid for by a curvature $a$ strictly between $q$ and $1$ and by a large
enough turning angle $\theta$.

Now place the tuning on the edge $h=cX$, where its rate on the quadratic of curvature $q$ is about $1-c/\kappa$, and write
$1-\cos\theta=2Xw$ for some $w>0$. Suppose that $X$ is small and that $\eta$ is much smaller than $X$, which is the
case for the tunings that matter as $\kappa\to\infty$. Then~\eqref{eq:A} and~\eqref{eq:ReA}, with $a=\re A/h$ and $b=\im A/h$, give
$a\approx4w/c$, $|b|\approx2\sqrt{Xw}/c$ and $\tan(\theta/2)\approx\sqrt{Xw}$. After division by $\sqrt{Xw}/c$, the condition above
becomes $2\le4w(1-4w/c)$, that is,
\[
  c\,(2w-1)\ \ge\ 8w^2.
\]
Some $w$ satisfies this if and only if
\[
  c\ \ge\ \min_{w>1/2}\frac{8w^2}{2w-1}=8,
\]
and the minimum is attained at $w=1$. At $c=8$ the best spiral has $a=1/2$, the middle of the range of curvatures,
$m=1/8$ and $|b|=\sqrt X/4$, and it turns by $\theta\approx2\sqrt X$ per step. More precisely, with $\vartheta=\eta/X$,
the polynomial $\Phi$ of \cref{lem:spiral} satisfies
\[
  \Phi(cX,2Xw)=2X^2\bigl[c\,(2w-1+\vartheta)-8w^2\bigr]+O(X^3).
\]
A tuning that beats $1-8/\kappa$ on quadratics has $h>8X$, so it lies beyond the edge $c=8$, and decaying spirals at the
rate $1-8/\kappa$ exist for it. The proof of \cref{thm:lower} makes this precise. It finds a window of values of $(h,u)$
around $u=2X$ on which $\Phi>0$ (\cref{lem:window}), and a root of unity inside it (\cref{lem:rou}).

\subsection{The upper bound}
\label{sec:howupper}

For the converse we need a tuning that lies on the edge $h=CX$ for its own rate, with $C$ just below $8$, and a proof
that no function in $\F$ slows it down. The tuning~\eqref{eq:tuning} satisfies $X=s(4+Cs)$ and $h=CX$, so its effective
step is $C$. Its momentum agrees with Polyak's to first order, $1-\beta\approx4/\sqrt\kappa$, while its step
$\gamma L\approx32/\sqrt\kappa$ is about $\sqrt\kappa/8$ times smaller than Polyak's.

The proof uses a Lyapunov function. Let $\varphi=f-(\mu/2)\|\cdot\|^2$, which is convex and $(1-q)$-smooth. To
leading order in $s$, the Lyapunov function is
\[
  V_t\approx f(x_{t-1})+\frac{\|x_t-x_{t-1}\|^2}{2\gamma}+\frac12\langle x_t-x_{t-1},\nabla\varphi(x_{t-1})\rangle
  +2s\|\nabla\varphi(x_{t-1})\|^2.
\]
The first two terms are the potential and the kinetic energy of the heavy ball. The third couples the velocity with
the gradient, and the fourth is a small multiple of the squared gradient. We show that $V_{t+1}\le r^2V_t$ on every $f\in\F$. The proof adds two inequalities satisfied by
$\varphi$, one between consecutive iterates and one between the iterate and the minimizer, with multipliers $r^2$ and
$1-r^2$. These are the only multipliers that cancel the function values. What remains is a quadratic form in
$x_t$, $x_{t-1}$, $\nabla\varphi(x_{t-1})$ and $\nabla\varphi(x_t)$ that must be positive semidefinite. Because the
quadratic $(\mu/2)\|x\|^2$ attains the rate $r$, the form must vanish in the corresponding direction, and this fixes
three of the six entries of the matrix that defines $V$. We found the other three numerically, as power series in $s$,
rounded them, and proved positivity for all $\kappa\ge400$ by exact root counting. \Cref{sec:why8} predicts this
outcome. With effective step $C<8$ no spiral can slow the tuning down, and the certificate shows that no other function
can either.

\section{Proof of the Lower Bound}
\label{sec:lower}

This section proves \cref{thm:lower}. Fix $q<1/8$, so that $r=1-8q\in(0,1)$. For $q\ge1/8$ there is nothing to prove.
Write $\eta=1-r=8q$. The proof has four steps. A tuning that beats $r$ on quadratics lies in a triangle
(\cref{lem:jury}). A decaying spiral at rate $r$ is a heavy-ball trajectory on some function of $\F$ as soon as a
polynomial $\Phi$ is non-negative (\cref{lem:spiral}). This polynomial is positive on an explicit window
(\cref{lem:window}), and some root of unity falls in that window (\cref{lem:rou}).

\subsection{Tunings that beat \texorpdfstring{$1-8/\kappa$}{1-8/kappa} on quadratics}

For a tuning $(h,\beta)$ define
\begin{equation}
  W=\frac{r^2-\beta}{r},\qquad X=W+\eta=1-\frac\beta r,\qquad\text{so that}\qquad\beta=r(1-X).
  \label{eq:WX}
\end{equation}

\begin{lemma}[quadratic necessary conditions]
  \label{lem:jury}
  If $\rho_Q<r$, then
  \[
    \eta<X<\frac{2(2-\eta)}{10-\eta}\le\frac25\qquad\text{and}\qquad 8X<h<(2-\eta)(2-X).
  \]
  In particular $0<\beta<1$, $W>0$ and $h>0$.
\end{lemma}

\begin{proof}
  By the Schur--Cohn criterion, both roots of $z^2-az+b$, with $a$ and $b$ real, lie in the open disc of radius $r$ if
  and only if $|b|<r^2$ and $|a|r<r^2+b$. We apply it to $\lambda=q$, where $a=1+\beta-qh$, and to $\lambda=1$, where
  $a=1+\beta-h$. First, $\beta<r^2$ gives $W>0$, that is, $X>\eta$. With $\beta=r(1-X)$ one checks that
  \[
    (r^2+\beta)-(1+\beta-qh)r=r\,(qh-\eta X)
    \qquad\text{and}\qquad
    (r^2+\beta)+(1+\beta-h)r=r\,\bigl((2-\eta)(2-X)-h\bigr).
  \]
  So $\lambda=q$ gives $qh>\eta X$, that is, $h>8X$, and $\lambda=1$ gives $h<(2-\eta)(2-X)$. The two are compatible only
  if $8X<(2-\eta)(2-X)$, that is, $X<2(2-\eta)/(10-\eta)$, which is at most $2/5$. Then $\beta=r(1-X)\in(0,1)$.
\end{proof}

These conditions describe the interior of the sublevel set $\mathrm{SLS}(r)$ of \citet[Lemma~2.4]{goujaud2025provable}
in the coordinates $(X,h)$. The edge $h=8X$ is their edge $\gamma=(1-\rho)(1-\beta/\rho)/\mu$, the edge $X=\eta$ is
$\beta=\rho^2$, and the third edge is the stability edge of the stiff direction. The compatibility condition forces
$\eta<6-4\sqrt2$, so for $q\ge0.043$ every tuning has $\rho_Q\ge r$.

\subsection{Decaying spirals}

Identify $\R^2$ with $\C$, with the inner product $\langle v,v'\rangle=\re(v\bar v')$. For an integer $K\ge3$ let
$\theta=2\pi/K$, $z=re^{i\theta}$ and $u=1-\cos\theta=2\sin^2(\pi/K)$. If heavy-ball is to follow $x_t=z^t$,
then~\eqref{eq:hb} forces the gradients
\begin{equation}
  h\,g_t=(1+\beta)x_t-\beta x_{t-1}-x_{t+1}=A\,x_t,\qquad\text{where}\qquad
  A=1+\beta-\frac{\beta}{r}e^{-i\theta}-re^{i\theta}.
  \label{eq:A}
\end{equation}
With $\beta=r^2-rW$ a short computation gives
\begin{equation}
  \re A=\alpha:=u(2-\eta-X)+\eta X\qquad\text{and}\qquad\im A=-W\sin\theta.
  \label{eq:ReA}
\end{equation}

\begin{lemma}[decaying spirals]
  \label{lem:spiral}
  Let $r\in(0,1]$, $\eta=1-r$ and $K\ge3$, with $\theta$, $z$ and $u$ as above. Let $(h,\beta)$ be a tuning with $h>0$
  and $W\ge0$, where $W$ and $X$ are defined by~\eqref{eq:WX} with this $r$, let $\alpha$ be as in~\eqref{eq:ReA}, and
  let
  \begin{equation}
    \Phi(h,u)=-qh^2+h\bigl[(1+q)\alpha-(1-q)(2-u)W\bigr]-\alpha^2-W^2u(2-u).
    \label{eq:Phi}
  \end{equation}
  If $\Phi(h,u)\ge0$, there is an $f\in\F(\R^2)$ on which heavy-ball started at $(x_{-1},x_0)=(z^{-1},1)$ gives
  $x_t=z^t$ for all $t\ge0$. In particular $\rho_{\gamma,\beta}(\F(\R^2))\ge r$.
\end{lemma}

\begin{proof}
  Take the data $x_t=z^t$, $g_t=(a+ib)x_t$ and $f_t=(a/2)|x_t|^2$ for $t\ge0$, with $a=\re A/h$ and $b=\im A/h$,
  together with the minimizer $(0,0,0)$. By~\eqref{eq:A}, heavy-ball maps $(x_{t-1},x_t)$ to $x_{t+1}$ on any function
  that interpolates these data. Substituting into~\eqref{eq:thg} gives, for all $i\ne j$,
  \begin{equation}
    I(i,j)=m\,|x_i-x_j|^2+b\,\im(x_j\bar x_i),\qquad I(i,\star)=I(\star,i)=m\,|x_i|^2,
    \label{eq:reduction}
  \end{equation}
  where $m=\bigl((a-q)(1-a)-b^2\bigr)/\bigl(2(1-q)\bigr)$ is the quantity of \cref{sec:why8}.
  For $\{i,j\}=\{n,n+k\}$ with $k\ge1$, this is $r^{2n}\bigl[m|1-z^k|^2\pm b\,r^k\sin k\theta\bigr]$, and both signs
  occur. So the data are interpolable if and only if $m\ge0$ and
  \begin{equation}
    m\,|1-z^k|^2\ \ge\ |b|\,r^k|\sin k\theta|\qquad\text{for all }k\ge1.
    \label{eq:allk}
  \end{equation}
  Since
  \[
    |1-z^k|^2=(1-r^k)^2+2r^k(1-\cos k\theta)\ge4r^k\sin^2\frac{\pi k}K
    \qquad\text{and}\qquad
    |\sin k\theta|=2\,\Bigl|\sin\frac{\pi k}K\cos\frac{\pi k}K\Bigr|,
  \]
  the right side of~\eqref{eq:allk} vanishes when $\sin(\pi k/K)=0$. For the other $k$, inequality~\eqref{eq:allk}
  follows from $2m|\sin(\pi k/K)|\ge|b|\,|\cos(\pi k/K)|$. Since $|\tan(\pi k/K)|\ge\tan(\pi/K)$ for these $k$, this
  holds as soon as $|b|\le2m\tan(\pi/K)$, a condition that implies $m\ge0$. Now $h|b|=W\sin\theta$ by~\eqref{eq:ReA}, and
  $\sin\theta/\tan(\pi/K)=1+\cos\theta=2-u$. Multiplying by $(1-q)h^2/\tan(\pi/K)>0$ turns the condition into
  \[
    2(1-q)h^2m-(1-q)hW(2-u)\ge0.
  \]
  Expanding $2(1-q)h^2m=(\alpha-qh)(h-\alpha)-(\im A)^2$ with $(\im A)^2=W^2u(2-u)$ shows that this is the condition
  $\Phi(h,u)\ge0$, and this condition implies $m\ge0$. So every finite subset of the data is interpolable, and
  \cref{lem:infinite} below provides a single $f\in\F(\R^2)$.
\end{proof}

At $r=1$ the bound $|1-z^k|^2\ge4r^k\sin^2(\pi k/K)$ is an equality, the condition $|b|\le2m\tan(\pi/K)$ is necessary as
well as sufficient, and among the tunings that converge on quadratics, the set $\{\Phi\ge0\}$ is the roots-of-unity
cycling region of \citet[Thm.~3.5]{goujaud2025provable} (checked symbolically). \Cref{lem:spiral} is therefore the
decaying version of their cycles, and the single-frequency case of the decaying waves of
\citet[Lemma~3.4, Thm.~3.5(c)]{pasechniuk2026two}. What it adds is a closed-form condition, $\Phi\ge0$, that can be
checked for all tunings at once.

\begin{lemma}[infinite interpolation]
  \label{lem:infinite}
  Let $(x_t,g_t,f_t)_{t\ge0}$ in $\R^d\times\R^d\times\R$ be such that every finite subfamily, together with
  $(x_\star,0,f_\star)$, is $\F$-interpolable. Then a single $f\in\F(\R^d)$ has $f(x_t)=f_t$ and $\nabla f(x_t)=g_t$ for
  all $t$, and $f(x_\star)=f_\star$ and $\nabla f(x_\star)=0$.
\end{lemma}

\begin{proof}
  Let $f_N\in\F(\R^d)$ interpolate the first $N$ points and $(x_\star,0,f_\star)$. Then
  $f_\star+(\mu/2)\|x-x_\star\|^2\le f_N(x)\le f_\star+(L/2)\|x-x_\star\|^2$, and the gradients $\nabla f_N$ are
  $L$-Lipschitz with $\nabla f_N(x_\star)=0$. By the Arzel\`a--Ascoli theorem and a diagonal argument, a subsequence has
  $f_N\to f$ and $\nabla f_N\to G$ locally uniformly. Then $f$ is $C^1$ with $\nabla f=G$. Moreover,
  $f-(\mu/2)\|\cdot\|^2$ is convex as a pointwise limit of convex functions, and $\nabla f$ is $L$-Lipschitz as a
  limit of $L$-Lipschitz maps. So $f\in\F(\R^d)$. Since $f_N$ interpolates the point $t$ for every $N>t$, and
  $(x_\star,0,f_\star)$ for every $N$, so does $f$.
\end{proof}

\Citet[Thm.~3.2]{pasechniuk2026two} prove the interpolation of infinite data sets directly, and it follows from the
extension theorem of \citet{azagra2017extension} as well.

\subsection{The window}

From now on we set $W=X-\eta$ and $q=\eta/8$ in~\eqref{eq:Phi}, so that $\Phi$ is a polynomial in $h$, $u$, $\eta$ and
$X$.

\begin{lemma}[window]
  \label{lem:window}
  Let $0<\eta<X\le2/5$. Then $\Phi(h,u)>0$ for all
  \[
    h\in\bigl[8X,\,(2-\eta)(2-X)\bigr]\qquad\text{and}\qquad
    u\in\bigl[2X(1-\sqrt X/2),\,2X(1+\sqrt X/2)\bigr].
  \]
\end{lemma}

\begin{proof}
  The polynomial $\Phi$ is concave in $h$, since the coefficient of $h^2$ is $-q$, and concave in $u$, since the
  coefficient of $u^2$ is $W^2-(2-\eta-X)^2=-4(1-\eta)(1-X)<0$, independently of $h$. A function on a rectangle that is concave in each
  variable separately attains its minimum at a corner, so it suffices to check the four corners. Write $u=2X(1+\nu)$
  with $\nu=\pm\sqrt X/2$, so that $4\nu^2=X$, and write $X=y^2$ and $\eta=\vartheta X$ with
  $(\vartheta,y)\in(0,1)\times(0,\sqrt{2/5}\,]$. The following polynomial identities are checked symbolically. At the
  lower corner $h=8X$,
  \begin{equation}
    \Phi(8X,u)=X^2\bigl[4X(X+2+7\nu)+2\vartheta E_1+2\vartheta^2XE_2\bigr]=y^4F_1(\vartheta,y),
    \label{eq:corner1}
  \end{equation}
  where $E_1=8-5X+2X\nu+8X\nu^2-8X^2(1+\nu)^2$ and $E_2=2X(1+\nu)-2\nu-3$. At the upper corner $h=(2-\eta)(2-X)$,
  \begin{equation}
    \Phi\bigl((2-\eta)(2-X),u\bigr)=\bigl(1-X(1+\nu)\bigr)\,y^2F_2(\vartheta,y),
    \label{eq:corner2}
  \end{equation}
  where
  \[
    F_2=4\bigl(2+4\nu-X(3+4\nu)\bigr)+\vartheta\bigl(6-9X+11X^2-16X\nu(1-X)\bigr)-\frac{\vartheta^2X}{2}\bigl(6+3X-X^2\bigr).
  \]
  Here $1-X(1+\nu)\ge1-0.4\cdot1.32>0$. Both $F_1$ and $F_2$ are quadratics in $\vartheta$ with negative leading
  coefficients, namely $2y^2E_2$, where $E_2\le2X(1+|\nu|)+2|\nu|-3<0$, and $-(y^2/2)(6+3y^2-y^4)$. So they are
  concave in $\vartheta$, and it suffices to check $\vartheta\in\{0,1\}$. With $\nu=\pm y/2$ this leaves the univariate
  polynomials
  \begin{alignat*}{2}
    F_1(0,y)&=4y^2\bigl(y^2\pm7y/2+2\bigr), &\qquad F_1(1,y)&=16-8y^2\pm14y^3-4y^4\mp14y^5-4y^6,\\
    F_2(0,y)&=8\pm8y-12y^2\mp8y^3, & F_2(1,y)&=14\pm8y-24y^2\mp16y^3+19y^4/2\pm8y^5+y^6/2.
  \end{alignat*}
  Each of $F_1(0,y)/y^2$, $F_1(1,y)$, $F_2(0,y)$ and $F_2(1,y)$ is positive at $y=7/11$ and has no root in
  $[0,7/11]\supset[0,\sqrt{2/5}\,]$, by exact Sturm counts. For instance, $y^2-7y/2+2>0$ for $y<0.71$. Hence
  $F_1>0$ and $F_2>0$ on the box.
\end{proof}

\Cref{app:polys} gives an independent check of the same lemma by interval arithmetic.

\subsection{Roots of unity in the window}

\begin{lemma}[roots of unity]
  \label{lem:rou}
  For every $X\in(0,2/5]$ there is an integer $K\ge4$ with
  $2X(1-\sqrt X/2)\le u_K\le2X(1+\sqrt X/2)$, where $u_K=2\sin^2(\pi/K)$.
\end{lemma}

\begin{proof}
  Let $y=\sqrt X$, $u^+=2X(1+y/2)$ and $u^-=2X(1-y/2)$, and let $K$ be the least integer $K\ge3$ with $u_K\le u^+$. Since
  $u_3=3/2>1.06>u^+$, we have $K\ge4$ and $u_{K-1}>u^+$. As $\sin x/x$ decreases on $(0,\pi]$,
  $\sin(\pi/K)/\sin(\pi/(K-1))\ge(K-1)/K$, so $u_K\ge(1-1/K)^2u_{K-1}>(1-1/K)^2u^+$. From $u_K\le u^+$ and $\pi/K\le\pi/4$
  we get $1/K\le\arcsin(\zeta)/\pi$ with $\zeta=\sqrt{u^+/2}=y\sqrt{1+y/2}$. So $u_K\ge u^-$ follows from
  \begin{equation}
    \Bigl(1-\frac{\arcsin\zeta}{\pi}\Bigr)^2\ \ge\ \frac{u^-}{u^+}=\frac{1-y/2}{1+y/2},
    \label{eq:rou}
  \end{equation}
  which we now prove for all $y\in(0,\sqrt{2/5}\,]$.

  We first bound $\arcsin$ by a linear function. Since $y\le\sqrt{2/5}$,
  \[
    \zeta=y\sqrt{1+y/2}\ \le\ \sqrt{2/5}\,\sqrt{1+\sqrt{2/5}/2}\ <\ 0.7256\ <\ 3/4.
  \]
  Since $\arcsin$ is convex on $[0,1]$ and vanishes at $0$, $\arcsin\zeta\le(4\zeta/3)\arcsin(3/4)$ for $\zeta\in[0,3/4]$. The
  alternating Taylor series gives
  \[
    \sin0.8481\ \ge\ 0.8481-\frac{0.8481^3}{6}+\frac{0.8481^5}{120}-\frac{0.8481^7}{5040}\ >\ 0.75002,
  \]
  so $\arcsin(3/4)<0.8481$. With $\pi>3.1415$ this gives $\arcsin(\zeta)/\pi\le0.36\,\zeta$.

  Next we reduce~\eqref{eq:rou} to a polynomial inequality. Put $w=\sqrt{1+y/2}=1+\delta$, where
  \[
    0\ <\ \delta\ \le\ \sqrt{1+\sqrt{2/5}/2}-1\ <\ 0.1473.
  \]
  Then $y=2(w^2-1)$ and $0.36\,\zeta=0.72\,w(w^2-1)$. Since
  $u^-/u^+=(2-w^2)/w^2$, \eqref{eq:rou} follows from
  \[
    w\bigl(1-0.72\,w(w^2-1)\bigr)\ \ge\ \sqrt{2-w^2},
  \]
  both sides being non-negative.

  Finally, $w(w^2-1)=\delta(1+\delta)(2+\delta)\le2.4637\,\delta$ for $\delta\le0.1473$, and $0.72\cdot2.4637\le1.78$. So the left side
  is at least $(1+\delta)(1-1.78\delta)=1-0.78\delta-1.78\delta^2>0$. The right side is $\sqrt{1-2\delta-\delta^2}\le1-\delta-\delta^2/2$. The
  difference of the two bounds is $\delta(0.22-1.28\delta)\ge0$ for $\delta\le0.1718$.
\end{proof}

Each constant in this proof is checked in exact arithmetic by \file{code/verify_rou_analytic.py}. Independently,
\file{code/verify_lowerbound.py} checks~\eqref{eq:rou} on $[0.1,\sqrt{2/5}]$ by interval arithmetic on $4000$
subintervals.

\subsection{Proof of \texorpdfstring{\cref{thm:lower}}{Theorem A}}

If $\rho_Q\ge r$ there is nothing to prove, since $\rho_{\gamma,\beta}(\F)\ge\rho_Q$. Otherwise \cref{lem:jury} gives
$\eta<X\le2/5$ and $h\in(8X,(2-\eta)(2-X))$, with $W>0$. \Cref{lem:rou} gives a $K\ge4$ with $u_K$ in the window of
\cref{lem:window}, so $\Phi(h,u_K)>0$. \Cref{lem:spiral} then gives an $f\in\F(\R^2)$ with $\|x_t\|=r^t$. Extending $f$
by $(\mu/2)\|\cdot\|^2$ in the other coordinates gives the result in $\R^d$. Since
$\|x_T\|=r^{T+1}\max\{\|x_0\|,\|x_{-1}\|\}$, the bound~\eqref{eq:rate} fails for every $\rho<r$. \qed

\subsection{Rest start}
\label{sec:rest}

\begin{proof}[Proof of \cref{cor:rest}]
  We first treat quadratics. On $(\lambda/2)x^2$ started at rest, $x_1=(1-h\lambda)x_0$, and $x_t$ is a linear combination of
  $z_1^t$ and $z_2^t$ when the roots are distinct. If the coefficient of the dominant root $z_1$ vanishes, then $1-h\lambda=z_2$, so
  $z_1=\beta$, and either $\beta=0$, which is gradient descent with $x_t=(1-h\lambda)^tx_0$, or $z_2=1$, and the method
  does not converge. The case of a double root is similar. So heavy-ball started at rest still converges at the rate
  $r(\lambda)$ or does not converge, and $\rho^\rest_{\gamma,\beta}\ge\min\{\rho_Q,1\}$.

  Now suppose $\rho_Q<r$, and take $K$ and the planar spiral data of the proof of \cref{thm:lower}, in the plane $\C$.
  Replace the first gradient by $g_0=(1-z)/h$, so that heavy-ball started at rest at $x_0=1$ lands on $x_1=z$. Add a
  third coordinate, on which $f$ is $(\lambda/2)y^2$ for some $\lambda\in(q,1)$, started at $y_{-1}=y_0=Y$, so that the
  space is $\C\times\R\cong\R^3$. Then $y_t=Ya_t$
  with $a_{-1}=a_0=1$ and $a_{t+1}=(1+\beta-h\lambda)a_t-\beta a_{t-1}$. The inequalities~\eqref{eq:thg} split over
  orthogonal blocks,
  \[
    I(i,j)=I^\circ(i,j)+c_\lambda(y_i-y_j)^2,\qquad c_\lambda=\frac{(\lambda-q)(1-\lambda)}{2(1-q)}>0,
  \]
  with $y_\star=0$, where $I^\circ$ is computed from the planar data. For $i,j\ge1$ and the minimizer, $I^\circ\ge0$ by
  \cref{lem:spiral}. The pairs that involve the index $0$ have $|I^\circ|\le M_0$ for a constant $M_0$, because all data
  are bounded ($r<1$). Each $a_j(\lambda)-1$ with $j\ge1$ is a polynomial of degree $j$ in $\lambda$ with leading coefficient
  $(-h)^j\ne0$. So for all but countably many $\lambda$, $a_j\ne1$ for every $j\ge1$, and since $a_j\to0$, this gives
  $\delta_\lambda=\inf_{j\ge1}(1-a_j)^2>0$. Taking $Y^2\ge M_0/(c_\lambda\min\{\delta_\lambda,1\})$ makes every inequality hold.
  \Cref{lem:infinite} gives an $f\in\F(\R^3)$, and $\|x_t\|\ge|z^t|=r^t$ with $x_{-1}=x_0=(1,Y)$.
\end{proof}

This is the construction of \citet[§5.1 and Lemma~A.5]{pasechniuk2026two}, specialized to one decaying wave.

\section{Proof of the Upper Bound}
\label{sec:upper}

This section proves \cref{thm:upper}. Throughout, $L=1$, $\ell=1-q$, $x_\star=0$ and $f(x_\star)=0$.

\subsection{The tuning on quadratics}

With the tuning~\eqref{eq:tuning}, $1+\beta-hq=r+\beta/r$ is an identity in $s$, so the polynomial
$z^2-(1+\beta-hq)z+\beta$ of the direction of curvature $q$ has the roots $r$ and $\beta/r=r-4s$. Equivalently,
$X=1-\beta/r=s(4+Cs)$ and $h=CX$, so the tuning lies on the edge $h=CX$ of \cref{sec:quadratics} for its own rate.
Running heavy-ball on $(\mu/2)\|x\|^2$ from $x_{-1}=x_0/r$ gives $x_t=r^tx_0$. Started at rest, both roots are excited,
and by the argument in the proof of \cref{cor:rest} the coefficient of $r^t$ is non-zero. In the direction of curvature
$1$ the roots are complex with modulus $\sqrt\beta<r$, because the discriminant $(1+\beta-h)^2-4\beta$ is negative for
$0<s\le1/20$. So $\rho_Q=r$, and it remains to show that no function in $\F$ does worse than the quadratics.

\subsection{The Lyapunov function}

Let $\varphi=f-(q/2)\|\cdot\|^2$. It is convex and $\ell$-smooth, with $\varphi(0)=0$ and $\nabla\varphi(0)=0$. Let
$p_t=\nabla\varphi(x_t)$, so that $\nabla f(x_t)=p_t+qx_t$ and heavy-ball reads
\begin{equation}
  x_{t+1}=(1+\beta-hq)x_t-\beta x_{t-1}-h\,p_t.
  \label{eq:hbphi}
\end{equation}
Convexity and smoothness of $\varphi$ give, for consecutive iterates and for the minimizer,
\begin{align*}
  I^1_t&=\varphi(x_{t-1})-\varphi(x_t)-\langle p_t,x_{t-1}-x_t\rangle-\frac1{2\ell}\|p_{t-1}-p_t\|^2\ \ge\ 0,\\
  I^2_t&=-\varphi(x_t)+\langle p_t,x_t\rangle-\frac1{2\ell}\|p_t\|^2\ \ge\ 0.
\end{align*}
With $\xi_t=(x_t,x_{t-1},p_{t-1})$ and a symmetric $3\times3$ matrix $P=(P_{ij})$, define
\begin{equation}
  V_t=\xi_t^\top(P\otimes I_d)\,\xi_t+\varphi(x_{t-1}).
  \label{eq:V}
\end{equation}

\begin{proposition}[dissipation identity]
  \label{prop:identity}
  Let $P=P(s)$ be the matrix of \cref{app:P}. Along~\eqref{eq:hbphi},
  \begin{equation}
    r^2V_t-V_{t+1}-r^2I^1_t-(1-r^2)I^2_t=v_t^\top(F\otimes I_d)\,v_t,\qquad\text{where}\qquad
    v_t=(x_t,x_{t-1},p_{t-1},p_t),
    \label{eq:identity}
  \end{equation}
  for a symmetric $4\times4$ matrix $F=F(s)$ of rational functions of $s$. For $0<s\le1/20$, $F\succeq0$ and
  $G=P+e_3e_3^\top/(2\ell)\succ0$.
\end{proposition}

The values of $\varphi$ cancel in~\eqref{eq:identity}. The multiplier $r^2$ of $I^1_t$ removes the term
$r^2\varphi(x_{t-1})$ of $r^2V_t$, and since the two multipliers add up to $1$ they remove the term $-\varphi(x_t)$ of
$-V_{t+1}$. The identity is between quadratic forms with scalar coefficients, so it holds in
every dimension once it holds for $d=1$.

\begin{proof}[Proof of \cref{thm:upper} from \cref{prop:identity}]
  By~\eqref{eq:identity}, $V_{t+1}\le r^2V_t-r^2I^1_t-(1-r^2)I^2_t\le r^2V_t$, so $V_t\le r^{2t}V_0$. Since
  $\varphi(x)\ge\|\nabla\varphi(x)\|^2/(2\ell)$, we have $V_t\ge\xi_t^\top(G\otimes I_d)\xi_t\ge\lambda_{\min}(G)\|x_t\|^2$.
  Moreover $\|p_{-1}\|\le\ell\|x_{-1}\|$ and $\varphi(x_{-1})\le(\ell/2)\|x_{-1}\|^2$, so
  \[
    V_0\le\|P\|\bigl(\|x_0\|^2+(1+\ell^2)\|x_{-1}\|^2\bigr)+\frac\ell2\|x_{-1}\|^2,
  \]
  where $\|P\|$ is the spectral norm.
  Hence $\|x_t\|\le Mr^t(\|x_0\|+\|x_{-1}\|)$ with
  \[
    M^2=\frac{\|P\|(1+\ell^2)+\ell/2}{\lambda_{\min}(G)}.
  \]
  Together with the quadratic lower bound above, $\rho_{\gamma,\beta}=\rho^\rest_{\gamma,\beta}=r$.
\end{proof}

\subsection{Structure of the certificate}

On $(\mu/2)\|x\|^2$ we have $p\equiv0$, and the slow solution has $x_{t-1}=x_t/r$. All the $I$'s vanish there,
and $V$ decays at the rate $r^2$. So the right side of~\eqref{eq:identity} vanishes at $v_0=(1,1/r,0,0)$, and
$F\succeq0$ forces $Fv_0=0$. This determines $P_{11}$, $P_{12}$ and $P_{13}$ from the other entries (\cref{app:P}). It
makes $F\succeq0$ equivalent to $F_3\succeq0$, where $F_3$ is the lower-right $3\times3$ block of $F$. Indeed,
writing $v=v^{(1)}v_0+w$ with $w^{(1)}=0$ gives $v^\top Fv=w^\top Fw$. The three free entries are
\begin{align*}
  P_{22}&=\frac1s\Bigl(\frac1{64}+\frac{5}{32}s+\frac54s^2+26s^3\Bigr),\\
  P_{23}&=-\Bigl(\frac14+\frac98s+11s^2+248s^3\Bigr),\\
  P_{33}&=s\,\bigl(2+14s+441s^2+3210s^3\bigr).
\end{align*}
To leading order in $s$, using $\gamma\approx32s$ and $\varphi+(q/2)\|\cdot\|^2=f$,
\begin{equation}
  V_t=f(x_{t-1})+\frac{\|x_t-x_{t-1}\|^2}{2\gamma}+\frac12\langle x_t-x_{t-1},\nabla\varphi(x_{t-1})\rangle
  +2s\|\nabla\varphi(x_{t-1})\|^2+\cdots,
  \label{eq:energy}
\end{equation}
where the dots stand for the cross terms $(s/2)\langle x_t-x_{t-1},x_{t-1}\rangle$ and
$4s^2\langle x_{t-1},\nabla\varphi(x_{t-1})\rangle$ to leading order, and for corrections of higher order in $s$ to the
coefficients shown. The certificate is therefore a small perturbation of the mechanical energy of the heavy ball,
potential plus kinetic.

\subsection{Exact verification}
\label{sec:verification}

In the basis
\[
  U=\begin{pmatrix}1&4&0\\0&1&1\\0&-1&1\end{pmatrix},\qquad\det U=2,
\]
the leading principal minors of $U^\top F_3U$ and of $G$ are rational functions of $s$ with the expansions
\begin{alignat*}{4}
  &U^\top F_3U:&\qquad &\frac18+\frac34s+O(s^2),&\qquad &\frac s4+O(s^2),&\qquad &2s^3+O(s^4),\\
  &G:& &\frac1{64s}+O(1),& &\frac s{128}+O(s^2),& &\frac s{256}+O(s^2).
\end{alignat*}
Write each minor as $n(s)/d(s)$ in lowest terms. We verify by an exact Sturm count that the polynomial $nd$ has no root in $(0,1/20]$, and
that it is positive at $s=1/20$. So all six minors are positive on $(0,1/20]$, and Sylvester's criterion gives
$U^\top F_3U\succ0$ and $G\succ0$. The identity~\eqref{eq:identity}, the null vector, the root $r$ and the sign of the
discriminant in the direction of curvature $1$ are all checked symbolically or by exact root counting. This proves
\cref{prop:identity}, and with it \cref{thm:upper}.

\subsection{Exact range}
\label{sec:range}

The minors of $G$ and the first minor of $U^\top F_3U$ have no sign change on $(0,1/5)$. The second minor of
$U^\top F_3U$ first changes sign at $s\approx0.060896$ ($\kappa\approx269.67$) and the third at $s_1\approx0.0529041$, that is,
$\kappa=1/s_1^2\approx357.2901$. The discriminant for curvature $1$ stays negative up to $s\approx0.122$. At $s_1$ the
third minor vanishes and the other five are positive, so $F_3\succeq0$ by continuity and the conclusions of
\cref{prop:identity} still hold. For every $s\in(s_1,1)$, either $P$ is undefined or some minor is negative, so that
$F\not\succeq0$ or $G\not\succ0$. So this certificate proves \cref{thm:upper} for every $\kappa\ge1/s_1^2$ and for no
smaller $\kappa$. We state $\kappa\ge400$ for simplicity. For $64.82<\kappa<1/s_1^2$ the rate of the same tuning still
follows from \citet{pasechniuk2026two} (\cref{rem:pv}), but there $8-64/\sqrt\kappa$ is below $4.62$, while the
certified tuning at $\kappa=10^2$ in \cref{tab:bracket} already gives $5.032$.

\subsection{How the certificate was found}
\label{sec:found}

For fixed $s$, the matrices $P$ with $F\succeq0$ and $G\succ0$ form a spectrahedron, which we explored numerically with
a semidefinite programming solver. Imposing the null vector leaves three unknowns. At $s=1/50$, $1/100$ and $1/400$ its
extent in $P_{22}$ and in $P_{23}$ is at most about $12s$ times the size of these entries, so the set is thin, while
$P_{33}$ can vary by about three times its size and change sign. We computed its analytic center along the family
$C=8-\tau s$, whose rate is $1-8/\kappa+\tau\kappa^{-3/2}$, fitted the expansions of the three entries in powers of $s$, rounded the coefficients to simple rationals,
and verified the result exactly. None of this search is part of the proof. A formal expansion of the minors suggests
that, within this certificate class and the family $\beta=r(r-\sigma s)$, feasibility as $s\to0$ requires
$\tau>6\sigma+96/\sigma$, which is smallest at $\sigma=4$, where it reads $\tau>48$. So the value $\tau=64$ is not far from
what lag-one certificates of this form can give. We did not try to prove this threshold.

\section{Computer-Assisted Steps}
\label{sec:computer}

This section states what was checked by computer, how, and on what range. No step of either proof relies on
floating-point arithmetic.

\subsection{What was checked}

\Cref{tab:computer} lists every computer-assisted step of the proofs, together with the auxiliary exact computations
quoted in the text. ``Symbolic'' means an identity between polynomials or rational functions with rational
coefficients, verified by expanding the difference to zero. ``Sturm'' means an exact count of the real roots of a
univariate polynomial with rational coefficients on a closed interval, together with an exact evaluation of its sign at
one point. ``Interval'' means outward-rounded interval arithmetic.

\begin{table}[ht]
  \centering
  \caption{Computer-assisted steps. ``Range'' is the exact parameter range on which the check holds. The scripts are in
    \texttt{code/}.}
  \label{tab:computer}
  \footnotesize
  \setlength{\tabcolsep}{4pt}
  \begin{tabular}{>{\raggedright\arraybackslash}p{1.7cm}>{\raggedright\arraybackslash}p{4.1cm}>{\raggedright\arraybackslash}p{2.95cm}>{\raggedright\arraybackslash}p{1.5cm}>{\raggedright\arraybackslash\scriptsize}p{3.5cm}}
    \toprule
    Claim & What is checked & Range & Method & Script \\
    \midrule
    \cref{lem:jury} & the two Schur--Cohn identities & all parameters & symbolic & \file{verify_lowerbound} \\
    \cref{lem:spiral} & reduction~\eqref{eq:reduction}, formulas~\eqref{eq:ReA}, identity for $\Phi$ & all parameters & symbolic & \file{verify_lowerbound} \\
    & at $r=1$, $\Phi$ gives the region of \citet[Thm.~3.5]{goujaud2025provable} & all parameters & symbolic & \file{verify_lowerbound}, \file{verify_symbolic} \\
    \cref{lem:window} & corner identities~\eqref{eq:corner1} and~\eqref{eq:corner2}, $u^2$-coefficient & all parameters & symbolic & \file{verify_lowerbound} \\
    & 8 end polynomials and 4 concavity coefficients have no root & $y\in[0,7/11]$ & Sturm & \file{verify_lowerbound} \\
    & (independent) $F_1,F_2>0$ on the box & $[0,1]\times[0,\sqrt{2/5}]$ & interval, 76 boxes & \file{verify_lowerbound} \\
    \cref{lem:rou} & the constants of the proof & as stated & exact rationals & \file{verify_rou_analytic} \\
    & (independent) inequality~\eqref{eq:rou} & $y\in[0.1,\sqrt{2/5}]$ & interval, 4000 pieces & \file{verify_lowerbound} \\
    \cref{prop:identity} & identity~\eqref{eq:identity}, null vector $v_0$, $r$ is a root for curvature $q$ & all $s$ & symbolic & \file{verify_achievability} \\
    & 6 leading minors positive & $s\in(0,1/20]$ & Sturm & \file{verify_achievability} \\
    & discriminant for curvature $1$ negative & $s\in(0,1/20]$ & Sturm & \file{verify_achievability} \\
    & first sign change of each minor, validity at $s_1$ and failure for every larger $s$ & $s\in(0,1)$ & real-root isolation & \file{verify_achievability} \texttt{-{}-range} \\
    \cref{rem:pv} & membership in the region $Q$ of \citet{pasechniuk2026two} (sufficient conditions) & $s\in(0,0.12420979]$ & Sturm & \file{verify_pv_membership} \\
    \cref{tab:bracket} & fixed-$\kappa$ certificates (\cref{app:fixed}) & $\kappa\in\{10^2,400,10^3,10^4\}$ & exact rationals & \file{fixed_kappa_certificates} \\
    \cref{sec:ghadimi} & spiral bound for the corner tuning & $\kappa=10^2,\dots,10^5$ & interval & \file{ghadimi_corner} \\
    \bottomrule
  \end{tabular}
\end{table}

\subsection{Size of the computations and trust base}

In \cref{lem:window} the polynomials have degree at most $6$ and small rational coefficients. They are printed in the
proof and in \cref{app:polys}, and can be checked by hand. In \cref{prop:identity} the six products $nd$ have degrees
$12$, $26$ and $34$ (minors of $U^\top F_3U$) and $14$, $18$ and $26$ (minors of $G$), with integer coefficients of up to
$21$ digits. The discriminant for curvature $1$ has degree $12$. The entries of $P$ are printed in \cref{app:P}. The tuning, the
matrices $P$, $F$, $U$ and $G$, the six minors with their products $nd$, and the discriminant are written in exact form to
\texttt{results/optimal/achievability.json}, so the check can be repeated in any computer algebra system.

All exact computations use SymPy~1.14 under Python~3.12, namely rational arithmetic, polynomial expansion,
\file{Poly.count_roots} (Sturm sequences) and \file{Poly.real_roots} (real-root isolation). Interval arithmetic uses
\texttt{mpmath}~1.3 (\texttt{mpmath.iv}, 30 to 40 digits, outward rounding). Each check runs in under a minute on a
laptop.

The proofs are not formalized in a proof assistant, so the trust base includes the correctness of the polynomial
routines of SymPy and of \texttt{mpmath.iv}. We reduce this dependence in three ways. The lower bound does not need
interval arithmetic, because \cref{lem:rou} has a proof by hand and the univariate polynomials of \cref{lem:window} are
short, and both lemmas are checked a second time by an independent route. The upper bound depends on exact root counts for
seven polynomials of degree at most $34$, which any computer algebra system can repeat from the published data. Finally, the
certificate of \cref{thm:upper} was found numerically, but only its exact form is used. Floating point appears only in
the numerical columns of \cref{tab:bracket} and in the tests below, and none of these is used in a proof.

\subsection{Floating-point tests}

As a sanity check, separate from the proofs, we built the objects of the proofs in floating point. For $400$ random
tunings with $q\in[10^{-7},0.042]$ that satisfy the conditions of \cref{lem:jury}, we built the spiral with the $K$ of
\cref{lem:rou} and checked~\eqref{eq:allk} directly for all $k\le3000$. The smallest relative slack was $0.15$. For $20$
random tunings with $\rho_Q<r$ and $q\in[10^{-4},0.04]$, we built the rest-start construction of \cref{cor:rest} in
$\R^3$, verified that heavy-ball started at rest reproduces it, and checked all pairwise inequalities~\eqref{eq:thg}
among up to $600$ points and the minimizer. The smallest slack was $4.9\cdot10^{-3}$.

\section{Finite Condition Numbers}
\label{sec:finite}

\Cref{cor:optimal} is asymptotic. This section compares it with what is known at finite $\kappa$.

\subsection{Exact certificates at fixed condition numbers}

The certificate class of \cref{sec:upper} applies to a single rational tuning at a fixed $\kappa$, where every entry
is a rational number. For $\kappa=10^2$, $400$, $10^3$ and $10^4$ we searched numerically for the fastest tuning in the
region $Q$ of \citet{pasechniuk2026two}, rounded it to a rational tuning whose rate $r$ is a root of the polynomial for
curvature $q$, solved a small semidefinite program for the free entries of $P$, rounded them, and verified the
certificate exactly (\cref{app:fixed}). This proves $c(\kappa)\ge5.032$, $6.173$, $6.734$ and $7.548$ at these values
of $\kappa$, in every dimension. These constants exceed $8-64/\sqrt\kappa$ at the same $\kappa$, so the tuning of
\cref{thm:upper} is not optimal at finite $\kappa$.

\subsection{A numerical bracket}

We computed two further quantities in floating point, which are not certified. On the side of tunings, we minimized the exact
rate $\rho_Q$ over a grid of tunings in the region $Q$ of \citet[Thm.~7.2(b)]{pasechniuk2026two}, where their theorem
gives $\rho=\rho_Q$ (\file{code/pv_region_search.py}). On the side of functions, for each tuning we took the larger of
$\rho_Q$ and the rate of the slowest decaying spiral, and minimized over a grid of tunings refined near the optimum.
The spirals are either root-of-unity spirals as in \cref{lem:spiral} or general log-spirals $x_t=r^te^{it\theta}$ with
any $\theta$, and for the latter the function values $f_t=Fr^{2t}$ may have any scale $F$. \Cref{tab:bracket} collects
these numbers with the proved bounds.

\begin{table}[ht]
  \centering
  \caption{Bounds on $c(\kappa)=\kappa(1-\rho^\star(\kappa))$ for $d\ge2$. Lower bounds on $c$ come from tunings, and
    upper bounds from functions that slow every tuning down. Lower bounds are rounded down and upper bounds up. The
    columns marked ``num.'' are floating point.}
  \label{tab:bracket}
  \small
  \begin{tabular}{rcccccc}
    \toprule
    & \multicolumn{3}{c}{lower bounds on $c$ (tunings)} & \multicolumn{3}{c}{upper bounds on $c$ (functions)} \\
    \cmidrule(lr){2-4}\cmidrule(lr){5-7}
    & & exact & best on $Q$ & log-spirals & root-of-unity & \\
    $\kappa$ & \cref{thm:upper} & (\cref{app:fixed}) & (num.) & (num.) & spirals (num.) & \cref{thm:lower} \\
    \midrule
    $10^2$ & -- & 5.032 & 5.034 & 5.37 & 5.44 & 8 \\
    $400$ & 4.800 & 6.173 & 6.175 & -- & 6.50 & 8 \\
    $10^3$ & 5.976 & 6.734 & 6.736 & 6.98 & 6.99 & 8 \\
    $10^4$ & 7.360 & 7.548 & 7.549 & -- & 7.66 & 8 \\
    $10^5$ & 7.797 & -- & 7.851 & -- & -- & 8 \\
    $10^6$ & 7.936 & -- & -- & -- & -- & 8 \\
    $\infty$ & 8 & & & & & 8 \\
    \bottomrule
  \end{tabular}
\end{table}

\subsection{Earlier numerical values}
\label{sec:earlier}

The thesis of \citet[Table~10.1]{goujaud2024thesis} reports numerical optima at $\kappa=20$, $100$ and $1000$. With
\cref{rem:conventions} they convert to $c\approx3.42$, $5.03$ and $7.49$, if they refer to the rate of the distance. The
value at $\kappa=100$ agrees with \cref{tab:bracket}, and the tuning listed with it, $\beta=0.65$ and $\gamma L=1.57$,
is close to our certified tuning $\beta=0.650$ and $\gamma L=1.587$ in \cref{tab:fixed}. We could not reconcile the
value at $\kappa=1000$. The tuning listed with it, $\beta=0.839$ and $\gamma=1.01$ read as $\gamma L$, already has a
rate on quadratics that corresponds to $c\approx6.5$, and $7.49$ exceeds our numerical spiral bound $6.98$. The
thesis may use a different notion of rate. \Citet{vanscoy2021speed} find $c\approx5$ at $\kappa=100$, in agreement with
\cref{tab:bracket}. The values of \citet[Fig.~8]{lessard2016analysis}, for the restricted family with step $1/L$, lie
below our numerical spiral bounds for $\kappa\lesssim4000$, as valid bounds must. Their values of $1-\rho$ appear to lie
on a grid of spacing about $2^{-12}$, which limits the resolution in $c$ to about $\kappa/4096$.

\subsection{Shape of the tuning and the second-order term}

Expanding~\eqref{eq:tuning} gives
\[
  \gamma L=32s-192s^2+O(s^3),\qquad 1-\beta=4s+16s^2+O(s^3),\qquad\frac{\gamma}{1-\beta}\to\frac8L.
\]
Polyak's tuning has $\gamma L=4/(1+s)^2$ and $1-\beta=4s/(1+s)^2$. The momenta agree to first order, and the step of
\cref{thm:upper} is smaller by a factor of about $\sqrt\kappa/8$. At $\kappa=10^4$ the tuning is $\gamma L=0.2998$ and $\beta=0.9586$,
with rate $r=1-7.36/\kappa$. The certified tunings of \cref{tab:fixed} have the same shape,
$\beta\approx r(r-\sigma s)$, with $\sigma=2.65$, $3.20$, $3.45$ and $3.85$ at $\kappa=10^2$, $400$, $10^3$ and $10^4$
against $\sigma=4$ in \cref{thm:upper}, so they use slightly more momentum.

\Cref{cor:optimal} gives $\rho^\star=1-8/\kappa+c_2\kappa^{-3/2}+\cdots$ with $c_2\in[0,64]$, if such an expansion
exists. At $\kappa=10^3$ and $10^4$ the numerical bracket gives values of $\sqrt\kappa\,(8-c(\kappa))$ between about $32$ and
$45$. This suggests that $c_2>0$, so that $\rho^\star(\kappa)$ is strictly above $1-8/\kappa$ at large finite
$\kappa$. Only the leading constant is proved.

\section{Discussion and Limitations}
\label{sec:discussion}

\subsection{What the results settle}

For $d\ge2$, the optimal worst-case rate of heavy-ball with constant parameters on $\F$ is $1-8/\kappa$ to first order,
and an explicit family of tunings attains it. On this class, momentum divides the number of iterations of gradient
descent by four asymptotically, whereas on quadratics it divides it by a factor of order $\sqrt\kappa$. In our family
the momentum is Polyak's to first order and the step is about $\sqrt\kappa/8$ times smaller, and the best tunings we found
numerically at fixed $\kappa$ have the same shape (\cref{sec:finite}). We have not proved that every asymptotically
optimal family must look like this.

\subsection{Open questions}

\Cref{thm:upper} is optimal only as $\kappa\to\infty$. At fixed $\kappa$ better tunings exist (\cref{sec:finite}), and
we have no closed form for them, so the problem posed by \citet[§10.3.3]{goujaud2024thesis} is solved to first order
only. The second-order term is open as well. We prove $c_2\in[0,64]$, while the numerics point to values between about $32$
and $45$. A second-order analysis of decaying spirals, possibly with several frequencies as in
\citet{pasechniuk2026two}, could raise the lower end. Lag-one certificates of our form seem unable to lower the upper
end below $48$ (\cref{sec:found}), so this may need certificates with longer memory. For $\kappa$ below $357.2901\ldots$
we have exact certificates at isolated values only.

The spirals need two dimensions. In one dimension \cref{thm:upper} still applies, so
$8-64/\sqrt\kappa\le c_1(\kappa)\le50/3+o(1)$ (\cref{app:onedim}), and whether $c_1(\kappa)\to8$ is open. A proof of
$c_1\le8+o(1)$ would require functions of one variable that slow every tuning down as much as the decaying spirals do,
and by \cref{prop:limit} single-jump cycles do not suffice. Similarly,
\cref{cor:rest} uses a third coordinate, and we do not know whether the rest-start rate on $\F(\R^2)$ obeys the same
lower bound. Our functions are not twice differentiable at the minimizer (\cref{sec:related}), whereas the
one-dimensional lower bound of \citet{liu2026sharp} holds for $C^\infty$ functions. Whether $1-8/\kappa$ persists on
smoother subclasses of $\F$ is open.

The results concern constant $(\gamma,\beta)$. Variable parameters and restarts are not covered, and neither are methods
that evaluate the gradient elsewhere, such as Nesterov's method and the triple momentum method, which accelerate. The
results are worst-case statements for exact gradients on smooth strongly convex functions, and they do not directly
describe the stochastic training of neural networks.

\subsection{The constant 8 in the region of Ghadimi et al.}
\label{sec:ghadimi}

\Citet[Lemma~2.5]{goujaud2025provable} show that the best rate on quadratics over the region where
\citet[Thm.~4]{ghadimi2015global} prove global convergence on $\F$ is $1-8/\kappa+o(1/\kappa)$. The two constants are
related but describe different tunings. For small steps, the region of Ghadimi et al.\ is $\gamma L\lesssim4(1-\beta)$.
In the triangle of \cref{sec:quadratics} for the rate $1-c/\kappa$, the corner where the momentum edge $X=\eta$ meets the
edge $h=cX$ of the effective step is $\beta=r^2$, $\gamma L=c^2/\kappa$, with $1-\beta\approx2c/\kappa$. It lies in the
region of Ghadimi et al.\ if and only if $c\lesssim8$. In floating point the minimum of their Lemma~2.5 is attained at
this corner, with $\gamma L=52.3/\kappa$, $62.6/\kappa$ and $63.9/\kappa$ at $\kappa=10^2$, $10^3$ and $10^4$. The step
printed in the lemma, $2(1+\beta^\star)/(L+\mu)$, exceeds $2/L$ and lies outside the region. Our tuning has
$\gamma L\approx32/\sqrt\kappa$ instead. On $\F$ the corner $\beta=(1-8/\kappa)^2$, $\gamma L=64/\kappa$ is much slower
than on quadratics. Decaying spirals (\cref{lem:spiral}, which holds for every $r$) prove $c\le3.681$, $3.972$, $3.998$
and $4.000$ for it at $\kappa=10^2$, $10^3$, $10^4$ and $10^5$ (interval arithmetic, \file{code/ghadimi_corner.py}). The
lag-one criterion of \citet{pasechniuk2026two} gives $c\ge2.92$, $3.14$, $3.16$ and $3.16$ there in floating point.
Their Lemma~2.5 therefore does not give $8$ on $\F$, and we know of no deeper connection between the two constants.

\subsection{Limitations}

The proofs of \cref{lem:window} and \cref{prop:identity} use exact computer algebra (\cref{sec:computer}). The numerical
results at finite $\kappa$ are in floating point and partly rely on a recent result that has not yet been refereed
\citep{pasechniuk2026two}, but none of them is used in a proof. Our literature search, up to October 15, 2026, found no
earlier statement of the limit $8$. Given the concurrent activity on this problem, independent work on the same question
is possible.

\section{Conclusion}
\label{sec:conclusion}

With constant parameters and the best tuning, heavy-ball converges on $L$-smooth $\mu$-strongly convex functions at the
rate $1-8/\kappa+O(\kappa^{-3/2})$ in every dimension $d\ge2$, four times faster than gradient descent. The lower bound
is a decaying root-of-unity spiral, which exists for every tuning that is faster than $1-8/\kappa$ on quadratics. The
upper bound is an explicit tuning, with the momentum of Polyak's tuning to first order and a step about
$\sqrt\kappa/8$ times smaller, certified by a
Lyapunov function close to the mechanical energy and verified exactly for all $\kappa\ge400$. The constant $8$ comes
from a trade-off between rotation and curvature. Along a spiral the gradient must rotate, and convexity allows rotation
only in exchange for curvature. This exchange is cheapest at half the largest curvature. The optimal tuning at finite $\kappa$, the
second-order term and the constant in one dimension remain open. The same pair of tools, decaying spirals and lag-one
certificates, may give the optimal constants of other methods with a fixed structure.

\gdef\sectionpage{\gdef\sectionpage{\clearpage}}
\section*{Use of AI}

This work was carried out by the AI system Claude Opus 5.5 (Anthropic), which produced the mathematical arguments,
the programs and the writing under the direction of the author.

\renewcommand{\bibsection}{\section*{\refname}\phantomsection\addcontentsline{toc}{section}{\refname}\markboth{\refname}{}}
\renewcommand{\bibfont}{\small\raggedright}
\setlength{\bibsep}{0.22\baselineskip} 
\bibliography{references}

\appendix
\addtocontents{toc}{\protect\setcounter{tocdepth}{1}}  

\section{The Certificate of Theorem~B}
\label{app:P}

As polynomials in $s=\kappa^{-1/2}$, the tuning of \cref{thm:upper} is
\begin{align*}
  r&=1-8s^2+64s^3,\\
  \gamma L&=32s-192s^2-1024s^3+4096s^4,\\
  \beta&=1-4s-16s^2+160s^3-192s^4-1024s^5+4096s^6.
\end{align*}
\Cref{tab:Mconst} gives the tuning and the constant $M$ of \cref{thm:upper} at a few values of $\kappa$.

\begin{table}[ht]
  \centering
  \caption{The tuning and the constant of \cref{thm:upper} at sample values of $\kappa$, rounded, with $M$ rounded
    up.}
  \label{tab:Mconst}
  \small
  \begin{tabular}{rcccccc}
    \toprule
    $\kappa$ & $r$ & $\beta$ & $\gamma L$ & $\|P\|$ & $\lambda_{\min}(G)$ & $M$ \\
    \midrule
    $400$ & $0.988$ & $0.778544$ & $1.0176$ & $1.44$ & $5.6\cdot10^{-4}$ & $78$ \\
    $10^3$ & $0.994024$ & $0.862348$ & $0.791643$ & $1.57$ & $2.3\cdot10^{-4}$ & $127$ \\
    $10^4$ & $0.999264$ & $0.958558$ & $0.299817$ & $3.51$ & $2.4\cdot10^{-5}$ & $559$ \\
    $10^6$ & $0.999992$ & $0.995984$ & $0.031807$ & $31.6$ & $2.5\cdot10^{-7}$ & $1.6\cdot10^4$ \\
    \bottomrule
  \end{tabular}
\end{table}

The Lyapunov matrix $P$ in~\eqref{eq:V} has the entries
\begin{align*}
  P_{11}&=\frac{25600s^5-1920s^4-1632s^3-100s^2-12s-1}{64s\,(16s^2-2s-1)\,r^2},\\
  P_{12}&=-\frac{26112s^5-1984s^4-1632s^3-92s^2-12s-1}{64s\,(16s^2-2s-1)\,r},\\
  P_{13}&=\frac{1984s^3+88s^2+9s+2}{8r},\\
  P_{22}&=\frac{1664s^3+80s^2+10s+1}{64s},\\
  P_{23}&=-\frac{1984s^3+88s^2+9s+2}{8},\\
  P_{33}&=s\,(3210s^3+441s^2+14s+2).
\end{align*}
The first three entries solve $Fv_0=0$ with $v_0=(1,1/r,0,0)$. In terms of the free entries, with $C=8-64s$ and
$\sigma=4$ the coefficient of $s$ in $\beta=r(r-\sigma s)$, they are
\begin{align*}
  P_{11}&=\frac{2CP_{22}s-2Cs^3+2\sigma P_{22}-\sigma s^2+2s}{2(Cs+\sigma)(Cs^2-1)^2},\\
  P_{12}&=\frac{2CP_{22}s-Cs^3+2\sigma P_{22}-\sigma s^2+s}{2(Cs+\sigma)(Cs^2-1)},\\
  P_{13}&=\frac{P_{23}}{Cs^2-1}.
\end{align*}
To leading order, $F=w_0w_0^\top/8+O(s)$ with $w_0=(1,-1,2,-2)$. The leading part of $F$ has rank one, and the positivity of
$F$ is decided at orders $s$ and $s^3$. The basis $U$ of \cref{sec:verification} makes this visible in the minors. In the
variables $e=x_t-x_{t-1}$, $x_{t-1}$ and $p=p_{t-1}$, the quadratic part of $V_t$ has the coefficients
\begin{alignat*}{3}
  \|e\|^2&:\ \frac1{64s}+\frac5{32}+O(s),&\qquad 2\langle e,x_{t-1}\rangle&:\ \frac s4+O(s^3),&\qquad
  \|x_{t-1}\|^2&:\ \frac{s^2}2+O(s^3),\\
  2\langle e,p\rangle&:\ \frac14+O(s),& 2\langle x_{t-1},p\rangle&:\ 2s^2+O(s^3),& \|p\|^2&:\ 2s+O(s^2).
\end{alignat*}
Together with $\varphi(x_{t-1})$ this gives~\eqref{eq:energy}.

\section{Polynomials of the Window Lemma}
\label{app:polys}

With $\nu=\pm y/2$, $X=y^2$ and $\eta=\vartheta y^2$, the corner polynomials of \cref{lem:window} are, sorted by powers of
$\vartheta$,
\begin{align*}
  F_1^{\pm}&=4y^2\bigl(y^2\pm7y/2+2\bigr)+2\vartheta\bigl(8-5y^2\pm y^3-6y^4\mp8y^5-2y^6\bigr)
             +2\vartheta^2y^2\bigl(-3\mp y+2y^2\pm y^3\bigr),\\
  F_2^{\pm}&=\bigl(8\pm8y-12y^2\mp8y^3\bigr)+\vartheta\bigl(6-9y^2\mp8y^3+11y^4\pm8y^5\bigr)
             -\frac{\vartheta^2y^2}2\bigl(6+3y^2-y^4\bigr).
\end{align*}
The second route proves $F_1^\pm>0$ and $F_2^\pm>0$ on $[0,1]\times[0,\sqrt{2/5}]$ by interval evaluation of the
monomial expansion and bisection. Since $F_1$ vanishes at $\vartheta=y=0$, the script writes
${F_1=F_1(0,y)+\vartheta F_{1a}}$ and shows that $F_1(0,y)/y^2$ and $F_{1a}$ are positive. This takes $9$ boxes
for $\nu>0$ and $67$ for $\nu<0$.

\section{Exact Certificates at Fixed Condition Numbers}
\sectionmark{Exact Certificates} 
\label{app:fixed}

Each row of \cref{tab:fixed} is a rational tuning with $r=1-c/\kappa$ a root of its polynomial for curvature $q$,
together with the free entries $(P_{22},P_{23},P_{33})$ of a certificate of the form of \cref{sec:upper}. The entries
$P_{11}$, $P_{12}$ and $P_{13}$ follow from $Fv_0=0$. The script \file{code/fixed_kappa_certificates.py} checks in exact
rational arithmetic that $r$ is the dominant root on quadratics and that the leading principal minors of $F_3$ and $G$
are positive, and hence that the rate of the tuning on $\F(\R^d)$ is $r$ in every dimension.

\begin{table}[ht]
  \centering
  \caption{Exact certificates at fixed $\kappa$ (\texttt{results/optimal/fixed\_kappa.json}).}
  \label{tab:fixed}
  \small
  \renewcommand{\arraystretch}{2.6}
  \setlength{\tabcolsep}{5pt}
  \begin{tabular}{rcccccc}
    \toprule
    $\kappa$ & $c$ & $\beta$ & $\gamma L$ & $P_{22}$ & $P_{23}$ & $P_{33}$ \\
    \midrule
    $100$ & $\dfrac{629}{125}$ & $\dfrac{203195907}{312500000}$ & $\dfrac{4958407}{3125000}$ & $\dfrac{368289}{10^6}$ & $-\dfrac{82409}{250000}$ & $\dfrac{40071}{250000}$ \\
    $400$ & $\dfrac{6173}{1000}$ & $\dfrac{101480295257}{125\cdot10^9}$ & $\dfrac{133279034566039}{123070937500000}$ & $\dfrac{521519}{10^6}$ & $-\dfrac{146987}{500000}$ & $\dfrac{4403}{50000}$ \\
    $1000$ & $\dfrac{3367}{500}$ & $\dfrac{878213437313}{10^{12}}$ & $\dfrac{387381978567129}{496633000000000}$ & $\dfrac{351169}{500000}$ & $-\dfrac{139493}{500000}$ & $\dfrac{14499}{250000}$ \\
    $10^4$ & $\dfrac{1887}{250}$ & $\dfrac{960020029523}{10^{12}}$ & $\dfrac{74017896690099}{249811300000000}$ & $\dfrac{349997}{200000}$ & $-\dfrac{51973}{200000}$ & $\dfrac{393}{20000}$ \\[4pt]
    \bottomrule
  \end{tabular}
\end{table}

\section{Results in One Dimension}
\label{app:onedim}

This appendix collects results on $\F(\R)$. Whether heavy-ball accelerates on functions of one variable, the first of
the two riddles of \citet{goujaud2025riddles} (their Open Problem~1), was settled negatively before this work. It was
first answered by \citet{fotopoulos2026cycling}, in a blog post of July 7, 2026, with a Lean~4 formalization of the core
construction that rests on two stated external inputs. They rotate the roots-of-unity cycles of
\citet{goujaud2025provable} by a suitable phase and project them onto a line. The blog reports, by personal
communication, that Dieuleveut and collaborators had also solved the problem. We found no such preprint as of October 15,
2026. Combined with \citet[Thm.~3.6]{goujaud2025provable}, this gives $\rho\ge1-(100/3+o(1))/\kappa$ on $\F(\R)$ for
every tuning. Building on that construction, \citet[App.~F, Thm.~8]{liu2026sharp} gave a complete proof with a fixed
phase. It shows that every tuning with $\gamma>0$ and $0\le\beta<1$ has rate at least $(\kappa-12)/(\kappa+12)$ for
$\kappa\ge24$, on $C^\infty$ functions with any Hessian-H\"older bound and with locally attracting cycles. \Citet{pasechniuk2026two}
credit the blog with settling the riddle. We include only what we believe adds to these works, namely a better constant
(\cref{thm:onedim}), the limit of the cycle method (\cref{prop:limit}), and an exact rest-start orbit of Polyak's tuning
for all $\kappa\ge100$ (\cref{thm:polyak-rest}).

\subsection{Projected rotations}

In one dimension, interpolation is a condition on consecutive secants. For $x_1<\dots<x_n$, some $f\in\F(\R)$ has
$f'(x_i)=g_i$ for all $i$ if and only if
\[
  \mu\ \le\ \frac{g_{i+1}-g_i}{x_{i+1}-x_i}\ \le\ L\qquad\text{for }i<n.
\]
Necessity is clear, and for
sufficiency $f'$ can be taken piecewise linear through the points $(x_i,g_i)$, with slope $\mu$ outside $[x_1,x_n]$.
No function values need to be prescribed, since heavy-ball uses only gradients, and coinciding points need equal
gradients. We use the following statement, due in this form
to \citet{fotopoulos2026cycling} for the interval of phases and to \citet[App.~F]{liu2026sharp} for the fixed phase. We
include a short proof for completeness.

Fix $K\ge3$, $\beta<1$ and $\gamma>0$, take $L=1$ so that $h=\gamma$ and $\gamma\mu=qh$, and let $\omega=2\pi/K$ and
$u=1-\cos\omega$. Define
\begin{equation}
  \alpha=(1+\beta)\,u,\qquad E=(1-\beta)\sin\omega>0,\qquad
  \Phi_K=(\alpha-qh)(h-\alpha)-E^2-E(1-q)h\cot\frac{\pi}{K}.
  \label{eq:phiK}
\end{equation}
Substituting $E\cot(\pi/K)=(1-\beta)(2-u)$ and $E^2=(1-\beta)^2u(2-u)$ shows that $\Phi_K$ is the polynomial
$\Phi(h,u)$ of~\eqref{eq:Phi} at $r=1$, and $\alpha$ is the $\alpha$ of~\eqref{eq:ReA} at $r=1$. For a phase $\psi$, let
$x_t=\cos(\omega t+\psi)$, with indices mod $K$, and $\gamma g_t=(1+\beta)x_t-\beta x_{t-1}-x_{t+1}$. These are the only
gradients for which heavy-ball runs through the points $x_t$ periodically.

\begin{lemma}[projected rotations]
  \label{lem:rotations}
  For $\psi\in(0,\pi/K)$ the points $x_t$ are distinct, and their sorted consecutive secant slopes, multiplied by
  $\gamma$, take the two values $\sigma_-=\alpha-E\cot\psi$ and $\sigma_+=\alpha+E\cot(\pi/K-\psi)$. Hence the $x_t$ form
  a heavy-ball $K$-cycle on some $f\in\F(\R)$ if and only if $qh\le\sigma_-$ and $\sigma_+\le h$, and such a phase exists
  if and only if $\Phi_K\ge0$.
\end{lemma}

\begin{proof}
  By the addition formulas, $\gamma g_t=\alpha x_t+Ey_t$ with $y_t=\sin(\omega t+\psi)$. Sorting by $x=\cos\theta$ means
  sorting by the folded angle $|\theta|\in(0,\pi)$. The folded angles of the points alternate between the two sets
  $\psi+\omega\mathbb Z$ and $-\psi+\omega\mathbb Z$, with gaps alternating between $2\psi$ and $\omega-2\psi$. For an
  upper point $(\cos A,\sin A)$ and a lower point $(\cos B,-\sin B)$ the scaled secant is $\alpha-E\cot((A-B)/2)$,
  which gives $\sigma_-$ for $A-B=2\psi$ and $\sigma_+$ for $A-B=-(\omega-2\psi)$. With $a_1=(\alpha-qh)/E$ and
  $a_2=(h-\alpha)/E$, the two conditions read $\psi\ge\acot a_1$ and $\pi/K-\psi\ge\acot a_2$. They are compatible if
  and only if $\acot a_1+\acot a_2\le\pi/K$, with $\acot$ taking values in $(0,\pi)$. Since $a_1+a_2=(1-q)h/E>0$, if
  $a_1\le0$ or $a_2\le0$ then $a_1a_2\le0$, so both this condition and $\Phi_K\ge0$ fail. For $a_1,a_2>0$ the addition
  formula for the cotangent turns the condition into $E^2(a_1a_2-1)-E^2(a_1+a_2)\cot(\pi/K)=\Phi_K\ge0$.
\end{proof}

The proof applies to every phase. Shifting the time shifts the phase by $\omega$, and the two slopes depend on the
phase only through twice the phase modulo $\omega$. So a phase $\psi'\in(-\pi/K,0)$ gives the same slopes as the phase
$\psi'+\pi/K\in(0,\pi/K)$, and a multiple of $\pi/K$ makes two points with different gradients coincide. Replacing $\psi$
by $\psi-\pi/K$ and reflecting $x\mapsto-x$, which adds $\pi$ to the phase, therefore keeps both slopes. The cycle then
visits its points in the order
\[
  (x_0,x_1,x_2,\dots,x_{K-2},x_{K-1})=\bigl(x^{(0)},x^{(1)},x^{(3)},x^{(5)},\dots,x^{(4)},x^{(2)}\bigr),
\]
where $x^{(0)}<x^{(1)}<\dots<x^{(K-1)}$ are the sorted points. This is the ordering conjectured by
\citet{goujaud2025riddles}, and we checked it for $K=3,\dots,39$. If $\Phi_K>0$ and $r(\mu)<1$,
flattening $f'$ to slope $\mu$ near each point of the cycle makes the cycle attracting, and mollifying makes $f$ smooth.
This is the one-dimensional version of \citet[§§4--5]{goujaud2025provable} and \citet[Thm.~8]{liu2026sharp}.

\subsection{The constant \texorpdfstring{$50/3$}{50/3}}

\begin{lemma}[explicit cycle region]
  \label{lem:window1d}
  Let $0<q<6/25$, $0\le\beta<1$, $\delta=1-\beta$, $0<h\le4$ and $h(6-25q)\ge50\delta$. Then $\Phi_K\ge0$ for some
  $K\ge4$.
\end{lemma}

\begin{proof}
  Write $u_K=1-\cos(2\pi/K)$. With $\beta=1-\delta$, \eqref{eq:phiK} becomes
  \begin{equation}
    \Phi_K=-\chi(u_K)+2qh\bigl(\delta+u_K(1-\delta)\bigr)+2u_K\delta(2u_K-\delta),
    \label{eq:phi-expansion}
  \end{equation}
  where $\chi(u)=4u^2-2hu+2h\delta+qh^2$.
  The middle term is non-negative, and so is the last one when $u_K\ge\delta/2$. It therefore suffices to find a $K\ge4$
  with $\chi(u_K)\le0$ and $u_K\ge\delta/2$.

  The quadratic $\chi$ is non-positive on $[u_-,u_+]$, where $u_-+u_+=h/2$ and $u_-u_+=(2h\delta+qh^2)/4$. Since
  $(u_-+u_+)^2/(u_-u_+)=h/(2\delta+qh)$ and $R+2+1/R=25/6$ at $R=3/2$, we get $u_+/u_-\ge3/2$ if and only if
  $h(6-25q)\ge50\delta$. The window lies at the right place, because $u_-\le h/4\le1=u_4$ and
  $u_-=u_-u_+/u_+>(2h\delta+qh^2)/(2h)>\delta$. The values $u_K$ decrease to $0$ with $u_K/u_{K+1}\le3/2$ for
  $K\ge4$. Indeed, $u_4/u_5=1.447\ldots$, and for $K\ge5$, $u_K/u_{K+1}\le((K+1)/K)^2\le1.44$ because $\sin x/x$
  decreases. Let $K^\ast=\max\{K\ge4:u_K\ge u_-\}$. Then $u_-\le u_{K^\ast}\le1.5\,u_{K^\ast+1}<1.5\,u_-\le u_+$ and
  $u_{K^\ast}>\delta/2$.
\end{proof}

\begin{lemma}[quadratic rate]
  \label{lem:quad}
  For $\beta\in[0,1)$ and $\gamma>0$, with $\bar s=1-\sqrt\beta$, we have $1-r(\mu)\le\min\{\bar s,\gamma\mu/\bar s\}$.
\end{lemma}

\begin{proof}
  Let $p(z)=z^2-(1+\beta-\gamma\mu)z+\beta$, so that $p(1)=\gamma\mu$. If the roots are complex, then $r=\sqrt\beta$, so
  $1-r=\bar s$, and $|1+\beta-\gamma\mu|<2\sqrt\beta$ gives $\gamma\mu>\bar s^2$. If they are real, $r_1\ge r_2$, then
  $r_1r_2=\beta\ge0$. If $r(\mu)\ge1$ there is nothing to prove. Otherwise $r\ge\sqrt\beta$, so $1-r\le\bar s$. If
  $0\le r_2\le r_1$, then $r_2\le\sqrt\beta$ and $1-r_1=\gamma\mu/(1-r_2)\le\gamma\mu/\bar s$. If $r_2\le r_1\le0$, then
  $\gamma\mu=(1-r_1)(1-r_2)\ge1\ge1-r$.
\end{proof}

\begin{theorem}[non-acceleration on $\F(\R)$ with constant $50/3$]
  \label{thm:onedim}
  For every $0<\mu<L$ and every $(\gamma,\beta)\in\R^2$,
  \[
    \rho_{\gamma,\beta}(\F(\R))\ \ge\ \frac{1-10q}{1+10q},
    \qquad\text{and, if }q\le1/25,\qquad
    \rho_{\gamma,\beta}(\F(\R))\ \ge\ 1-\frac{100q}{6+25q}.
  \]
  So $c_1(\kappa)\le100/(6+25/\kappa)\to50/3$. Moreover $\rho_{\gamma,\beta}\ge1$ if $\beta\in[0,1)$, $q<6/25$ and
  $h(6-25q)\ge50(1-\beta)$.
\end{theorem}

\begin{proof}
  We show that $1-\rho\le20q/(1+10q)$, and $1-\rho\le100q/(6+25q)$ when $q\le1/25$. The two bounds agree at $q=1/25$, and
  the first is trivial for $q\ge1/10$. A non-constant periodic orbit forces $\rho\ge1$, and $\rho\ge\rho_Q$ always. We go
  through the tunings case by case.

  If the tuning is unstable on quadratics, that is, $|\beta|\ge1$, $h\le0$ or $h\ge2(1+\beta)$, then
  $\rho\ge\rho_Q\ge1$. If $-1<\beta<0$ and $0<h<2(1+\beta)$, the roots of $p$ satisfy $r_1>0>r_2$, so
  $1-r_1=\gamma\mu/(1-r_2)<qh<2q$, and both target bounds exceed $2q$.

  Now let $0\le\beta<1$, $\delta=1-\beta$ and $0<h<2(1+\beta)$, so that in particular $h<4$, and first assume
  $q\le1/25$. If $h(6-25q)\ge50\delta$, then \cref{lem:window1d,lem:rotations} give a cycle, so $\rho\ge1$. Otherwise
  $qh<50q\delta/(6-25q)$ with $\delta=\bar s(2-\bar s)$ and $\bar s=1-\sqrt\beta$, and \cref{lem:quad} gives
  $1-\rho\le\min\{\bar s,\,50q(2-\bar s)/(6-25q)\}$. This is at most the crossing value
  $\bar s^\ast=100q/(6+25q)$, which is at most $20q/(1+10q)$ if and only if $q\le1/25$. If instead $1/25<q<1/10$,
  \cref{lem:quad} with $h<4$ gives $1-\rho\le\min\{\bar s,4q/\bar s\}\le2\sqrt q$, and $2\sqrt q\le20q/(1+10q)$ if and
  only if $10q-10\sqrt q+1\le0$, which holds for $(4-\sqrt{15})/10\le q\le(4+\sqrt{15})/10$, an interval that
  contains $(1/25,1/10)$.

  The last claim of the theorem follows from \cref{lem:window1d,lem:rotations} when $h\le4$. When $h>4$, we have
  $h>2(1+\beta)$, and the tuning is unstable on quadratics.
\end{proof}

The bound $c_1\le50/3+o(1)$ improves on $24$ \citep{liu2026sharp} and $100/3$
\citep{goujaud2025provable,fotopoulos2026cycling}. Together with \cref{thm:upper},
$8-64/\sqrt\kappa\le c_1(\kappa)\le50/3+o(1)$. In the notation of \cref{rem:conventions}, $50/3$ corresponds to
$C=25/3$, and should not be confused with the threshold $C>50/3$ of \citet{goujaud2025provable}.

\subsection{The limit of the cycle method}

We call a cycle on the $K$-th roots of unity \emph{single-jump} if it advances by one root per step, as the planar
cycles of \citet{goujaud2025provable} and their projections in \cref{lem:rotations} do. \Citet{fotopoulos2026cycling}
describe cycles that jump by several roots per step. Single-jump cycles combined with the quadratic bound cannot
prove anything below $16$.

\begin{proposition}[the cycle method stops at 16]
  \label{prop:limit}
  \leavevmode
  \begin{enumerate}[label=(\alph*)]
    \item If $0<\delta<1$, $h>0$ and $h(1+q)^2<8\delta(1-\delta)(1-q)$, then $\Phi_K<0$ for every $K\ge3$.
    \item At $\delta=32q$ and $h=8\delta(1-\delta)(1-q)^2/(1+q)^2$, which satisfies (a), we have $(1-\rho_Q)/q\to16$ as
          $q\to0$.
  \end{enumerate}
  So no combination of single-jump roots-of-unity cycles, with any $K$ and any phase, with the quadratic bound proves
  $\rho\ge1-c/\kappa$ with $c<16$ for all small $q$. This applies to the bounds of \citet{goujaud2025provable},
  \citet{fotopoulos2026cycling}, \citet{liu2026sharp} and \cref{thm:onedim}.
\end{proposition}

\begin{proof}
  (a) Dropping the terms $-qh^2$, $-2qh\delta u_K$ and $-2u_K\delta^2$ from~\eqref{eq:phi-expansion} leaves
  $\Phi_K\le-4(1-\delta)u_K^2+2h(1+q)u_K-2h\delta(1-q)$, whose maximum over $u_K$ is
  \[
    \frac{h}{4(1-\delta)}\bigl[h(1+q)^2-8\delta(1-\delta)(1-q)\bigr]<0.
  \]

  (b) Write $h=8\delta c_q$ with $c_q=(1-\delta)(1-q)^2/(1+q)^2\to1$. At curvature $\mu$, the substitution $z=1-v$ turns
  the characteristic polynomial into $v^2-(\delta+qh)v+qh$, so
  \[
    1-r(\mu)=\frac{16q\,c_q}{1+8qc_q+\sqrt{(1+8qc_q)^2-c_q}}\to16q.
  \]
  At curvature $L$, for $q<0.028$ the roots are complex, so $1-r(L)=1-\sqrt{1-32q}\to16q$.
\end{proof}

Part (a) is consistent with the description of the union of the cycle regions over real $K$ in
\citet[eqs.~(10.22)--(10.23)]{goujaud2024thesis}, and \citet[§7.4]{pasechniuk2026two} show that the union of these
regions is the complement of their region $\Omega_1$, on which they prove convergence (their Thm.~7.1(i)). The proof of
(a) bounds $\Phi_K$ for every real $u_K\ge0$, and hence for every real $K>2$. In the description of
\citet[§7.4 and Thm.~7.1(ii)]{pasechniuk2026two}, every tuning with a cycle that jumps by several roots of unity has
$\Phi_K\ge0$ for some real $K>2$, so these cycles do not reach the tunings of (a) either. As far as we know
the value $16$ is new, though simple. In dimension two and higher, \cref{thm:lower} shows that the true constant, $8$,
is below this limit, which is why its proof uses decaying spirals instead of cycles.

\subsection{Polyak's tuning from rest}

Several works start heavy-ball at rest, $x_{-1}=x_0$: \citet{pasechniuk2026two} throughout, and
\citet[App.~B]{lessard2016analysis} and \citet[App.~H]{liu2026sharp} in their non-convergence examples.
Heavy-ball is invertible for $\beta\ne0$, so an orbit started at rest can land exactly on a cycle only if the cycle
contains a rest state, that is, $x_{K-1}=x_0$.

\begin{theorem}[Polyak's tuning fails from rest in one dimension]
  \label{thm:polyak-rest}
  Let $\kappa\ge100$, $L=1$, $s=\kappa^{-1/2}$, $\gamma=4/(1+s)^2$, $\beta=((1-s)/(1+s))^2$ and $\delta=1-\beta$. Let
  $f'$ be the piecewise-linear function through the points
  \[
    \Bigl(-1,\frac{-(1+\beta)+\delta/\beta}{\gamma}\Bigr),\quad
    \Bigl(-\frac\delta\beta,\frac{-\delta(1+2\beta)}{\beta\gamma}\Bigr),\quad
    \Bigl(0,\frac\delta\gamma\Bigr),\quad
    \Bigl(1,\frac1\gamma\Bigr),
  \]
  extended with slope $\mu$. Then $f\in\F(\R)$, and heavy-ball started at $x_{-1}=x_0=1$ produces the sequence
  $1,0,-1,-\delta/\beta,1,1,0,-1,\dots$ of period five. In particular it never converges.
\end{theorem}

\begin{proof}
  The points $(1,0,-1,-\delta/\beta,1)$ form a five-cycle with a repeated point, and the two gradients at the repeated
  point agree because $\beta(1+\delta/\beta)=1=x_0-x_1$. The order $-1<-\delta/\beta<0<1$ holds because
  $1-6s+s^2>0$. The six slope conditions $\gamma\mu\le\gamma\cdot\text{secant}\le\gamma L$ reduce to the positivity on
  $(0,1/10]$ of the rational functions of $s$ in \cref{tab:polyak-rest}. The binding one is $1-10s+2s^3-s^4$, which is
  decreasing and equals $0.0019$ at $s=1/10$. One-dimensional interpolation completes the proof.
\end{proof}

\begin{table}[ht]
  \centering
  \caption{The slope conditions of \cref{thm:polyak-rest}, with $\sigma$ the scaled secant $\gamma\cdot$secant and
    $s=\kappa^{-1/2}$. All entries are positive on $(0,1/10]$.}
  \label{tab:polyak-rest}
  \renewcommand{\arraystretch}{2.2}
  \begin{tabular}{lll}
    \toprule
    interval and $\sigma$ & $\sigma-\gamma\mu$ & $\gamma L-\sigma$ \\
    \midrule
    $[0,1]$, \ $\sigma=\beta$ & $\dfrac{1-3s}{1+s}$ & $\dfrac{3-s}{1+s}$ \\
    $[-\delta/\beta,0]$, \ $\sigma=1+3\beta$ & $\dfrac{4(1-s)}{(1+s)^2}$ & $\dfrac{4s(1-s)}{(1+s)^2}$ \\
    $[-1,-\delta/\beta]$, \ $\sigma=\dfrac{(1+\beta)(\beta-2\delta)}{\beta-\delta}$ &
      $\dfrac{2(1-10s+2s^3-s^4)}{(1+s)^2(1-6s+s^2)}$ & $\dfrac{2(1-2s+10s^3-s^4)}{(1+s)^2(1-6s+s^2)}$ \\[6pt]
    \midrule
    order of the points & $1-\delta/\beta=\dfrac{1-6s+s^2}{(1-s)^2}$ & $\delta/\beta=\dfrac{4s}{(1-s)^2}$ \\[6pt]
    \bottomrule
  \end{tabular}
\end{table}

Earlier results of this kind hold at a single $\kappa$ or in higher dimension.
\Citet[§4.6 and App.~B]{lessard2016analysis} show at $\kappa=25$, by a finite verification, that the rest start
$x_0=3.3$ enters the basin of an attracting three-cycle. \Citet[App.~H, Prop.~10]{liu2026sharp} show at $\kappa=100$ that
every rest start in $[37/50,19/25]$ converges to an attracting three-cycle. \Citet{pasechniuk2026two} prove failure from
rest for every $\kappa\ge9+4\sqrt5$, in $\R^3$ (their Thm.~2.2). \Cref{thm:polyak-rest} holds for all $\kappa\ge100$ in
one dimension, with exact landing on the orbit. Whether Polyak's tuning fails from rest in one dimension for every
$\kappa\ge9+4\sqrt5$ remains open. With a free middle point in place of $-1$, the same shape appears to be feasible for
$\kappa$ above about $45.76$. This threshold is numerical, and exact rational instances at $\kappa=49$ and $64$ are
verified.

\section{Reproducibility}
\label{app:repro}

All scripts are at \url{https://github.com/marcoshernanz/heavy-ball-optimal-rate}. They run with SymPy~1.14,
\texttt{mpmath}~1.3, NumPy and Matplotlib, and CVXPY~1.9.3 with Clarabel is needed only for the optional numerical
searches. \Cref{tab:repro} lists what each script checks or produces.

\begin{table}[ht]
  \centering
  \caption{Scripts in \texttt{code/} and what they check or produce.}
  \label{tab:repro}
  \small
  \begin{tabular}{>{\raggedright\arraybackslash}p{5.3cm}>{\raggedright\arraybackslash}p{9.2cm}}
    \toprule
    Script & Checks or produces \\
    \midrule
    \file{verify_lowerbound.py} & all checks for \cref{thm:lower} and \cref{cor:rest} in \cref{tab:computer} and the
      floating-point tests, writes \file{results/optimal/lowerbound.json} and ends with \texttt{ALL OK} \\
    \file{verify_rou_analytic.py} & the constants of the proof of \cref{lem:rou} \\
    \file{verify_achievability.py}~\texttt{1/20} & \cref{prop:identity} for $\kappa\ge400$, writes
      \file{results/optimal/achievability.json}; with \texttt{-{}-range} it checks the range of \cref{sec:range}, and
      with \texttt{-{}-table} it computes \cref{tab:Mconst} \\
    \file{verify_pv_membership.py} & the tuning~\eqref{eq:tuning} lies in the region $Q$ of \citet{pasechniuk2026two}
      (\cref{rem:pv}) \\
    \file{fixed_kappa_certificates.py} & \cref{app:fixed}, writes \file{results/optimal/fixed_kappa.json} \\
    \file{ghadimi_corner.py} & \cref{sec:ghadimi} \\
    \file{verify_symbolic.py}, \file{stallcycles.py} & the identities of \cref{app:onedim}, and the rest-start orbit of
      \cref{thm:polyak-rest} in exact rationals \\
    \file{pv_region_search.py}, \file{decaying_spirals.py}, \file{scan_spiral_constant.py} & the numerical columns of
      \cref{tab:bracket} (best on $Q$, root-of-unity spirals, log-spirals) \\
    \file{make_figures_optimal.py}~\texttt{paper} & \cref{fig:spiral} \\
    \bottomrule
  \end{tabular}
\end{table}

\end{document}

%% file: preamble.tex
\usepackage[T1]{fontenc}
\usepackage{amsmath, amssymb, mathtools}
\usepackage{stix2}
\usepackage[varqu, varl, scaled=0.95]{inconsolata}
\usepackage[english]{babel}
\usepackage{microtype}
\usepackage[a4paper, hmargin=2.75cm, vmargin=3cm, headheight=30pt, headsep=0.75cm, footskip=1.2cm, heightrounded]{geometry}

\usepackage{enumitem}
\setlist{itemsep=2pt, parsep=0pt, topsep=0pt}

\usepackage{xcolor}
\definecolor{accent}{rgb}{0.09, 0.5, 0.99}
\colorlet{quiet}{black!55}

\makeatletter
\g@addto@macro\normalsize{%
  \setlength{\abovedisplayskip}{10pt plus 2pt}%
  \setlength{\belowdisplayskip}{10pt plus 2pt}%
  \setlength{\abovedisplayshortskip}{\abovedisplayskip}%
  \setlength{\belowdisplayshortskip}{\belowdisplayskip}%
}
\makeatother

\usepackage{titlesec}
\titleformat{\section}{\normalfont\Large\bfseries}{\thesection}{1em}{}
\titleformat{\subsection}{\normalfont\large\bfseries}{\thesubsection}{1em}{}
\titleformat{\subsubsection}{\normalfont\normalsize\bfseries}{\thesubsubsection}{1em}{}
\titleformat{\paragraph}[runin]{\normalfont\normalsize\bfseries}{}{0pt}{}
\titlespacing*{\section}{0pt}{24pt plus 4pt minus 2pt}{9pt plus 1pt}
\titlespacing*{\subsection}{0pt}{17pt plus 3pt minus 2pt}{5pt plus 1pt}
\titlespacing*{\subsubsection}{0pt}{13pt plus 2pt minus 2pt}{3pt plus 1pt}
\titlespacing*{\paragraph}{0pt}{4pt plus 2pt}{0.6em}

\newcommand{\sectionpage}{\clearpage}
\AddToHook{cmd/section/before}{\sectionpage}
\AddToHook{cmd/appendix/after}{\clearpage{\LARGE\bfseries Appendix\par}\gdef\sectionpage{\gdef\sectionpage{\clearpage}}}

\usepackage{fancyhdr}
\makeatletter
\newcommand{\subtitle}[1]{\def\@subtitle{#1}}
\newcommand{\headnote}[1]{\def\@headnote{#1}}
\newcommand{\contact}[1]{\def\@contact{#1}}
\newcommand{\runningtitle}[1]{\def\@runningtitle{#1}}
\def\@runningtitle{\@title}
\subtitle{}
\headnote{}
\contact{}
\renewcommand{\sectionmark}[1]{\markboth{\thesection\quad #1}{}}
\fancypagestyle{first}{%
  \fancyhf{}%
  \fancyhead[L]{\small\@headnote}%
}
\makeatother

\AddToHook{cmd/tableofcontents/before}{\begingroup\setlength{\parskip}{0.2\baselineskip}\hypersetup{linkcolor=black}}
\AddToHook{cmd/tableofcontents/after}{\endgroup}

\usepackage{graphicx}
\graphicspath{{figures/}}
\usepackage{booktabs}
\usepackage{caption}
\renewcommand{\arraystretch}{1.15}

\usepackage[hang, flushmargin]{footmisc}
\usepackage[numbers, sort&compress]{natbib}
\renewcommand{\bibfont}{\small}
\newcommand{\doi}[1]{\href{https://doi.org/#1}{doi:#1}}

\usepackage[bookmarksnumbered, bookmarksopen, bookmarksdepth=3]{hyperref}
\hypersetup{colorlinks=true, allcolors=accent}
\makeatletter
\AtBeginDocument{\hypersetup{pdftitle={\@title\ifx\@subtitle\@empty\else: \@subtitle\fi}, pdfauthor={\@author}}}
\makeatother
\usepackage[capitalize, noabbrev]{cleveref}
\crefname{equation}{Eq.}{Eqs.}
\Crefname{equation}{Equation}{Equations}
\newcommand{\sectionsign}[1]{%
  \crefformat{#1}{\S##2##1##3}%
  \Crefformat{#1}{Section~##2##1##3}%
  \crefrangeformat{#1}{\S\S##3##1##4--##5##2##6}%
  \Crefrangeformat{#1}{Sections~##3##1##4--##5##2##6}%
  \crefmultiformat{#1}{\S\S##2##1##3}{ and~##2##1##3}{, ##2##1##3}{ and~##2##1##3}%
  \Crefmultiformat{#1}{Sections~##2##1##3}{ and~##2##1##3}{, ##2##1##3}{ and~##2##1##3}%
}
\sectionsign{section}
\sectionsign{subsection}
\sectionsign{subsubsection}
\AddToHook{cmd/appendix/after}{\crefalias{section}{appendix}\crefalias{subsection}{subappendix}} 

\makeatletter
\renewcommand{\maketitle}{%
  \thispagestyle{first}%
  \begingroup
  \setlength{\parskip}{0pt}\raggedright%
  \vspace*{0.8cm}%
  {\fontsize{22}{27}\selectfont\bfseries\@title\par}%
  \ifx\@subtitle\@empty\else
  \vspace{0.35cm}%
  {\fontsize{14}{19}\selectfont\@subtitle\par}%
  \fi
  \vspace{0.9cm}%
  {\large\@author\par}%
  \vspace{0.1cm}%
  {\small\ttfamily\color{quiet}\@contact\par}%
  \endgroup
}
\renewenvironment{abstract}{%
  \par\vspace{1cm}%
  {\large\bfseries\abstractname\par}%
  \noindent\ignorespaces%
}{%
  \par
}
\makeatother